\documentclass{article}

\usepackage[preprint,nonatbib]{neurips_2023}

\usepackage[utf8]{inputenc} 
\usepackage[T1]{fontenc}    
\usepackage{hyperref}       
\usepackage{url}            
\usepackage{booktabs}       
\usepackage{amsfonts}       
\usepackage{nicefrac}       
\usepackage{microtype}      
\usepackage{xcolor}         
\usepackage{graphicx}
\usepackage{enumitem}
\usepackage{comment}
\usepackage{multirow}
\usepackage{colortbl}
\usepackage{subfigure}

\usepackage{algorithm}
\usepackage{algorithmic}
\usepackage{amsmath,amsthm}
\usepackage{thmtools, thm-restate}
\theoremstyle{plain}
\newtheorem{theorem}{Theorem}

\newtheorem{lemma}{Lemma}
\newtheorem{corollary}[theorem]{Corollary}
\theoremstyle{definition}

\theoremstyle{remark}

\usepackage[capitalize,noabbrev]{cleveref}
\usepackage{mathtools}
\usepackage{bm}

\def\independent{\perp \!\!\! \perp}
\renewcommand{\P}{\mathbb{P}}
\newcommand{\p}{\mathrm{p}}
\newcommand{\E}{\mathbb{E}}
\newcommand{\R}{\mathbb{R}}
\newcommand{\vx}{\mathbf{x}}
\def\vr{{\bm{r}}}
\def\rvn{{\mathbf{n}}}
\newcommand{\rvx}{\mathbf{X}}
\newcommand{\that}{\hat{\bm{\theta}}}
\newcommand{\vtheta}{\bm{\theta}}
\newcommand{\rvr}{\mathbf{R}}
\newcommand{\rv}{\mathbf{r}}
\newcommand{\pa}{\mathrm{pa}}
\newcommand{\anc}{\mathrm{anc}}

\def\vn{{\bm{n}}}
\def\vb{{\bm{b}}}
\def\vz{{\bm{z}}}

\def\mB{{\bm{B}}}
\def\mD{{\bm{D}}}
\def\mR{{\bm{R}}}
\def\mI{{\bm{I}}}

\DeclareMathOperator*{\argmax}{arg\,max}
\DeclareMathOperator*{\argmin}{arg\,min}

\DeclarePairedDelimiterX{\Esub}[1]{\mathrm{E}}{\big]}{%
	\;\big[\;#1%
}
\DeclarePairedDelimiterX{\Ee}[2]{\mathrm{E}}{\big]}{%
	_{#1}\;\big[\;#2%
}

\DeclarePairedDelimiterX{\infdivx}[2]{(}{)}{%
	#1\;\delimsize\|\;#2%
}

\newcommand{\tr}{\vtheta_p}
\newcommand{\ta}{\vtheta_{\alpha}}
\newcommand{\ttil}{\tilde{\vtheta}}

\newcommand{\gp}{\mathcal{G}_{p}}
\newcommand{\ga}{\mathcal{G}_{\alpha}}
\DeclarePairedDelimiterX{\tat}[1]{\theta}{{}}{%
	^{#1}%
}
\newcommand{\vphi}{\bm{\phi}}
\newcommand{\data}{\mathcal{S}}

\newcommand{\defeq}{\vcentcolon=}

\newcommand{\fu}{\overline{\mathcal{V}}}
\newcommand{\nfu}{\mathcal{V}}

\newcommand{\T}{T} 
\newcommand{\real}{\mathbb{R}}
\newcommand{\inner}[2]{\langle #1, #2 \rangle}
\newcommand{\midx}{\mathcal{V}}
\newcommand{\namenn}{LAMM }
\newcommand{\prx}{\P_{\rvr \mid \rvx}}
\newcommand{\wa}{\mB^{\alpha}}
\newcommand{\sa}{\sigma^{\alpha}}

\newcommand{\jp}{\p_{\rvx}(\vx;\vtheta_p)}
\newcommand{\ja}{\p_{\rvx}(\vx;\ta)}
\newcommand{\fp}{ \prod_{j \neq s} \p_{\rvx_j \mid \rvx_{\pa_j}}(\vx_j \mid \vx_{\pa_j}; \vtheta_p)}
\newcommand{\fa}{ \prod_{j \neq s} \p_{\rvx_j \mid \rvx_{\pa_j}}(\vx_j \mid \vx_{\pa_j}; \ta)}

\newcommand{\cpd}{\p_{\rvx_s\mid \rvx_{\pa_s}}(\vx_s \mid \vx_{\pa_s}; \vtheta_p)}
\newcommand{\cpda}{\p_{\rvx_s\mid \rvx_{\pa_s}}(\vx_s \mid \vx_{\pa_s}; \ta)}

\title{Deception by Omission:\\ Using Adversarial Missingness to Poison Causal Structure Learning}

\author{Deniz Koyuncu \thanks{Corresponding author: koyund@rpi.edu} \\
Rensselaer Polytechnic Institute \\
Troy, NY 12180 \\ 
\And
Alex Gittens \\
Rensselaer Polytechnic Institute \\
Troy, NY 12180 \\ 
\And
B\"ulent Yener \thanks{This work was done in part while the author was visiting Google LLC, New York, NY, USA} \\
Rensselaer Polytechnic Institute \\
Troy, NY 12180 \\ 
\And
Moti Yung \\
Google LLC and Columbia University \\
New York, NY 10011 \\
}

\begin{document}

\maketitle

\begin{abstract}
  Inference of causal structures from observational data is a key component of causal machine learning; in practice, this data may be incompletely observed. Prior work has demonstrated that adversarial perturbations of completely observed training data may be used to force the learning of inaccurate causal structural models (SCMs). However, when the data can be audited for correctness (e.g., it is crytographically signed by its source), this adversarial mechanism is invalidated. This work introduces a novel attack methodology wherein the adversary deceptively omits a portion of the true training data to bias the learned causal structures in a desired manner. Theoretically sound attack mechanisms are derived for the case of arbitrary SCMs, and a sample-efficient learning-based heuristic is given for Gaussian SCMs. Experimental validation of these approaches on real and synthetic data sets demonstrates the effectiveness of adversarial missingness attacks at deceiving popular causal structure learning algorithms.
\end{abstract}

\section{Introduction and  Threat Model}
\label{sxn:intro}

The feasibility of controlling and compromising ML models through adversarial poisoning of the data sets used in training is well-established, and there exists a body of literature exploring both designing and defending against such attacks (see a recent survey \cite{dp-survey21}). In this work we consider causal structure learning under the setting of a novel adversarial model which we call {\em adversarial missingness} (AM). This adversarial model is introduced to highlight and explore the potential for adversaries to exploit the ubiquitous and mild phenomenon of missing data, which at times (e.g., when sample inputs are signed) is the only measure the adversary can employ~\cite{MisDat-survey21,multimputcomp18}.

Under the AM threat model the adversary can neither modify the data nor introduce adversarial samples.  Such a restriction arises, for example, when the authenticity and integrity of the data is ensured by cryptographic mechanisms such as sensors digitally signing their output records they contribute to the data provider. Modifying such records or introducing false records is intractable, as it entails the auxiliary task of producing a corresponding digital signature, which is intractable due to the unforgeability of digital signatures. The adversary is therefore limited to only being able to partially conceal {\em existing} data. Similarly, in longitudinal studies or medical records, adversarial perturbations are susceptible to detection by post-hoc auditing, so in this context adversarial missingness is an attractive attack model. 


The principals of the adversarial missingness model are: (i) an adversarial data provider, (ii) a modeler, and (iii) an optional data auditor. We assume that the adversary has access to  a large number of records drawn from the true causal model; its goal is to pass along some records to the modeler in their entirety and selectively withhold some portion of the remaining records in order to fool the modeler into learning an inaccurate structural causal model (SCM).

The goal of the modeler is to use the partially observed data supplied by the adversary to infer the causal structure that gave rise to the completely observed data set. The modeler may have access to an independent data auditor who can partially verify the correctness of the learned causal model. Such verification, for example, may consist of a guarantee that the observational distribution recovered from the causal discovery process has small KL-divergence from the observational distribution that generated the completely observed data set. This particular form of verification may be accomplished through the use of an independent supply of data, for instance.

 The adversarial pattern of missingness may be arbitrarily selected. The adversary may have a target adversarial SCM with causal graph $\mathcal{G}_\alpha$ to which it would like the modeler to converge, or its objective may simply be to degrade the accuracy of the learned structure. For example, the adversary may delete an edge in the true causal graph $\mathcal{G}_p$ to obtain $\mathcal{G}_\alpha$, and then implement an {\em adversarial missingness mechanism} to determine the subset of data to be withheld to induce the modeler to learn the causal structure given by the adversarial graph. 



\begin{figure}[t]
\centering
\includegraphics[width=0.6\columnwidth]{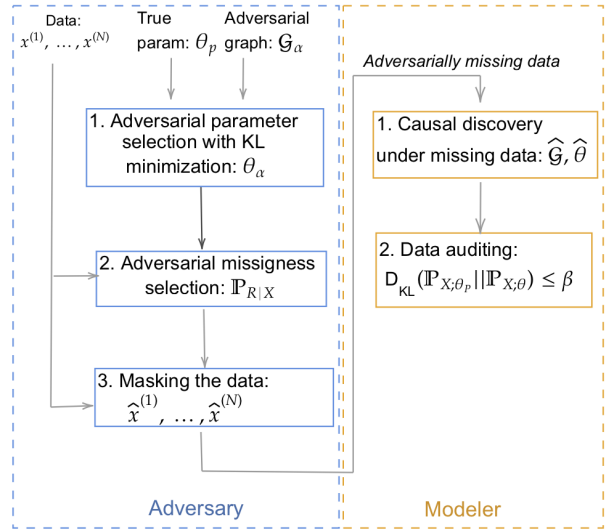}
\caption{The Adversarial Missingness (AM) threat model. The adversary masks a portion of the fully-observed data in order to induce the modeler to learn an SCM that is Markovian with respect to the desired DAG $\mathcal{G}_\alpha$. The modeler uses an auditor to ensure that the fitted SCM is plausible.}
\label{fig:overall}
\end{figure}

\paragraph{Contributions.} This work makes the following contributions: (i) it introduces a formulation of the AM model (Section~\ref{sxn:formulation}) that provides a proxy objective for arbitrary modelers and formalizes the objectives of the adversary, (ii) it uses rejection sampling to construct an AM attack with desirable theoretical properties (Section~\ref{sxn:generalized_rejection_sampling}), (iii) it introduces a neural parameterization of the adversarial missingness mechanism to design an AM attack heuristic which allows the adversary to trade-off between the missingness rate and the attack success (Section~\ref{sxn:neural_approach}), and (iv) it provides experimental validation on synthetic and real data to evaluate the performance of the two approaches to AM (Section~\ref{sec:exp}). This is an extended version of the conference paper~\cite{koyuncu}, with additional theoretical results characterizing the behaviors of the rejection sampling approaches and identifying optimal adversarial SCMs for both general distributions and Gaussian SCMs.

\paragraph{Notation.}
The structural causal models (SCMs) $\P_{\rvx}(\cdot;\vtheta)$ in this work are parameterized by parameter vectors $\vtheta$. The causal structure learning process ensures that these parameterized distributions have causal factorizations, so are valid SCMs. For convenience, the SCMs may be written with the parameters as subscripts, $\P_{\rvx;\vtheta}(\cdot)$. Similarly, the pdf of an SCM $\P_{\rvx;\vtheta}$ may be written as $\p_{\rvx;\vtheta}(\cdot)$ or $\p_\rvx(\cdot\,;\vtheta)$. The notation $\rvx \sim \P_{\rvx;\vtheta}$ indicates that $\rvx$ is a random vector governed by $\P_{\rvx;\vtheta}$. 

The true SCM underlying the fully observed data has parameter $\vtheta_p$ and is Markovian and faithful with respect to the true DAG $\mathcal{G}_p$. Adversarial SCMs have parameters $\vtheta_\alpha$ and are Markovian and faithful with respect to adversarial DAGs $\mathcal{G}_\alpha$. 

Each coordinate of the random vector $\rvr \in \{0,1\}^d$ indicates whether the corresponding entry of $\rvx$ is observed. Conditional distributions of the form $\P_{\rvr \mid \rvx}$, called missingness mechanisms, reflect how complete samples are used to determine which entries are observed.

The $i$th component of $\rvx$ is $\rvx_i$. The indices of the parents of $\rvx_i$ in its SCM are denoted by $\pa_i$. Given a subset of the indices $\mathcal{V}$, the corresponding subvector of $\rvx$ is denoted by $\rvx_\mathcal{V}.$ When an observation pattern $\vr$ is specified, the corresponding subvector of $\rvx$ that is observed is denoted by $\rvx_o$, and similarly the corresponding observed subvector of a fixed vector $\vx$ is denoted by $\vx_o$. The complementary unobserved subvectors are $\rvx_m$ and $\vx_m$.
\section{Background} 

 A \emph{causal factorization} of the distribution $\P$ of a random $d$-dimensional vector $\rvx$ explicitly identifies the causes of each variable $\rvx_i$, in the form
 $\P_\rvx = \prod_{i=1}^d \P_{\rvx_i \mid \rvx_{\pa_i} },$
 where $\pa_i$ denotes the indices of the parents of variable $\rvx_i$ in an associated directed acyclic graph (DAG) $\mathcal{G}$.  When $\P$ has a causal factorization corresponding to $\mathcal{G}$, it is \emph{Markovian} with respect to $\mathcal{G}$:  conditioned on its parents, each $\rvx_i$ is independent of its non-descendants. Conversely, $\P$ is said to be \emph{faithful} with respect to $\mathcal{G}$ if every conditional independence in $\P$ is encoded in $\mathcal{G}$.
 
 A structural causal model (SCM) associated with $\mathcal{G}$ expresses the cause-effect relationships using functional relations of the form $\rvx_i := f_i(\rvx_{\pa_i}, \rvn_i),$
 indicating that each variable is determined by the values of its parent variables, and a noise variable $\rvn_i$.
 Here, the exogenous noise variables $\rvn_1, \ldots, \rvn_d$ are assumed to be jointly independent of each other and the endogenous variables $\rvx$. 


In data-driven learning, including causal structure learning, missing data is commonly encountered. Missing data problems are studied under three basic models: (i) Missing Completely at Random (MCAR), (ii) Missing at Random (MAR), and (iii) Missing Not At Random (MNAR). In the MCAR model, the distribution of the missingness is independent of that of the features, while in the MAR model, the missingness depends at most on the observed features. In the most general model, MNAR, the missingness may depend on both the observed and unobserved features. Most algorithms with provable properties for dealing with missing data require the MCAR or MAR assumptions.

When missing data is present, standard causal structure learning algorithms cannot be used: direct application of structure learning algorithms using samples containing structured missingness may result in the learning of incorrect structures, e.g. due to selection bias. Instead, algorithms that learn structure from missing data must estimate the full data distribution from the incompletely observed data. Several recent approaches to causal structure learning have considered the presence of non-adversarial missing data~\cite{tu2019causal,thoemmes2015graphical,bhattacharya2020identification,10.5555/3020847.3020930,mohan2014graphical}. MissDAG~\cite{missdag} and MissGLasso~\cite{MissGLasso2010}, which motivate our assumptions on the modelers given in Section~\ref{sxn:formulation} and Section~\ref{sxn:neural_approach}, follow the same recipe for extending existing structure learning algorithms into the missing data setting: both propose to maximize the log-likelihood of the observed data assuming the missingness mechanism is ignorable. Next, both use the EM algorithm to maximize a lower bound iteratively. In MissGLasso, the maximization step uses the GraphLasso~\cite{graphlasso} penalty on the precision matrix and can be solved exactly. In MissDAG, the maximization step contains a DAG constraint (plus a sparsity penalty) and can only be solved approximately with, for example, the NOTEARS~\cite{notears} algorithm.



 A large body of work has arisen around data poisoning, presenting new attacks and defenses~\cite{dai2019backdoor,liu2018fine,fang2020influence} in applications ranging from text classification to image recognition models to recommendation systems. Attacks on causal discovery have only recently been investigated as an instance of data poisoning via insertion of adversarial samples~\cite{alsuwat2018cyber,alsuwat2020adversarial}. Both of these works consider the problem of adding data to the training set in order to influence the causal structures learned by the classical PC algorithm~\cite{spirtes2000causation}, and demonstrate the feasibility of both targeted and untargeted attacks. However, to our knowledge, no prior work has considered the use of missingness, rather than the insertion of false data, to manipulate the causal discovery process.

\section{Formulation of Adversarial Missingness}
\label{sxn:formulation}

The specific algorithm that the modeler uses to recover the SCM from the partially observed data is unknown. To mitigate this difficulty, we make reasonable assumptions on the modeler's structure learning algorithm to facilitate the design of practical adversarial missingness attacks; this process can be viewed analogously to the use of substitution attacks in standard adversarial ML to reduce attacks on models with unknown architectures into attacks on models with known architectures. Section~\ref{sec:exp} experimentally validates this approach by showing that attacks designed with these assumptions succeed even on structure learning algorithms that do not satisfy these assumptions.

Our two assumptions are: (i) the modeler assumes that the missingness mechanism is MAR, and (ii) the modeler seeks the causal structure that maximizes the probability of the partially observed data. The first assumption is motivated by the use of the MAR assumption in several approaches to learning causal structure from incompletely observed data. The second assumption is motivated by noting that in the case where the training data is completely observed, a common approach (e.g.~\cite{loh2014high,notears}) for the modeler is to learn a causal structure that maximizes the probability of the fully observed data subject to the distribution factorizing according to a DAG:
\begin{equation*}\label{eq:full}
	\that = \argmax_{\vtheta \in \mathcal{D} } \E_{\rvx; \vtheta_p}[ \log \P_{\rvx}(\vx;\vtheta) ].
\end{equation*}
Here, the distribution $\P_\mathbf{X}(\cdot; \vtheta)$ is from a parameterized family, and $\mathcal{D}$ is the set of parameters which satisfy the property that the corresponding distribution factorizes according to a DAG. 

When the training data is incompletely observed, under the MAR model, it is natural (e.g.~\cite{missdag,MissGLasso2010}) for the modeler to instead choose a causal structure that maximizes the probability of the partially observed data 
\begin{equation}\label{eqn:modeler-objective}
    \that = \argmax_{\vtheta \in \mathcal{D}} \E_{\rvr|\rvr\neq 0}\left[ \E_{\rvx_o|\rvr;\vtheta_p}\!\left[ \log \P_{\rvx_o;\vtheta}(\vx_o)\!\mid\! \rvr = \vr\right] \right].
\end{equation}

When the missingness model is indeed MAR, this formulation has the property that it leads to the same $\that$ as in the full-data case. 

\paragraph{Objective of the Modeler and Goals of the Adversary.} Equation~\ref{eqn:modeler-objective} is taken to be the objective of the modeler in our approach to AM. 
The solution $\that$ of \eqref{eqn:modeler-objective} is a function of the missingness mechanism $\P_{\rvr|\rvx}$. The adversary's aim is thus to find an adversarial SCM $\P_{\rvx;\vtheta_\alpha}$ and design an adversarial missingness mechanism $\P_{\rvr|\rvx}$ satisfying the following properties: (i) \textbf{adversarial Markovianity:} the adversarial SCM $\P_{\rvx;\vtheta_\alpha}$ is Markov relative to the adversarial graph $\mathcal{G}_\alpha$; (ii) \textbf{$\beta$-indistinguishability:} to foil the auditor, we impose the condition that the adversarial and true distributions must be within distance $\beta$ in KL-divergence; (iii) \textbf{bounded missingness rate:} the expected number of missing features per sample is bounded to reduce the chance that the modeler \textit{a priori} rejects the training data set as too incomplete to reliably infer causal structures; (iv) \textbf{attack success:} when \eqref{eqn:modeler-objective} is solved with $\P_{\rvr|\rvx}$, the distribution $\P_{\rvx;\that}$ learned by the modeler is Markov relative to $\mathcal{G}_\alpha$ and close in KL-divergence to $\P_{\rvx;\vtheta_\alpha}$.

 
 

The adversary's objective is thus to find an adversarial SCM parameterized by $\vtheta_\alpha$ and an adversarial missingness mechanism $\P_{\rvr|\rvx}$ that solve the constrained optimization problem
\begin{align}
\min_{\P_{\rvr|\rvx}, \vtheta_\alpha}\, & \text{D}_{\textrm{KL}}(\P_{\rvx;\vtheta_\alpha}\,\|\, \P_{\rvx;\that})  \label{eqn:adversarial-objective}\,\,\text{subject to}\,\, \begin{cases}
\text{D}_{\text{KL}}(\P_{\rvx;\vtheta_p}\,\|\, \P_{\rvx;\vtheta_\alpha}) \leq \beta, & \\
 \P_{\rvx;\vtheta_\alpha} \textrm{ is Markov relative to } \mathcal{G}_\alpha, \\
 \E_{\rvx;\vtheta_p} \left[ \E_{\rvr|\rvx} \left[\frac{|\{j\,|\,\rvr_j=0\}|}{d}\right] \right] \leq \gamma. 
\end{cases}
\end{align}
The attack success is measured by the KL-divergence between $\P_{\rvx;\that}$, the distribution learned by the modeler, and the target adversarial distribution $\P_{\rvx; \vtheta_\alpha}$, as well as the Structural Hamming distance between the DAG of the returned SCM and the target adversarial DAG $\mathcal{G}_\alpha$. This is an ambitious optimization problem, encoding multiple competing desiderata. 

\paragraph{A two stage approximation.} In the adversarial objective,~\eqref{eqn:adversarial-objective}, the adversary optimizes over the missingness mechanism and the adversarial SCM jointly, and $\beta$-indistinguishability is a difficult constraint to satisfy. We propose a two stage approximation to solving~\eqref{eqn:adversarial-objective}. 

In the first stage, the adversary selects a target adversarial SCM $\P_{\rvx; \ta}$ that is Markov with respect to $\mathcal{G}_\alpha$ and minimizes the KL-divergence to the true SCM. In the second stage, the adversary finds a missingness mechanism that guides the modeler to learn $\ta$ and has a bounded missingness rate. Specifically, given the target adversarial DAG $\mathcal{G}_\alpha$, we first solve
\begin{align}
	\ta=\argmin_{\vtheta \in \mathcal{D}_\alpha},  \text{D}_{\text{KL}}(\P_{\rvx;\vtheta_p}\,\|\, \P_{\rvx;\vtheta}) \label{eq:KLmin},
\end{align}
where $\mathcal{D}_\alpha\defeq\{\vtheta:\P_{\rvx;\vtheta} \textrm{ is Markov relative to } \mathcal{G}_\alpha\}$ denotes the set of feasible $\vtheta$. Next, given the adversarial SCM parameterized by $\vtheta_\alpha$, we relax the hard constraint in the original objective on the missingness rate using a Lagrange multiplier and solve for the missingness mechanism:
\begin{align}
	\min_{\P_{\rvr|\rvx}}\ \text{D}_{\textrm{KL}}(\P_{\rvx;\vtheta_\alpha}\,\|\, \P_{\rvx;\that}) + \lambda \E_{\rvx;\vtheta_p} \left[ \E_{\rvr|\rvx} \left[\frac{|\{j\,|\,\rvr_j=0\}|}{d}\right] \right]. \label{eqn:adversarial-objective2}
\end{align}
This two-stage approximation has the advantage of not requiring an a priori selection of $\beta$ and $\gamma$. Instead, the smallest possible $\beta$ is implicitly selected in the first stage, and by varying $\lambda$ the adversary can explore the trade-off between ensuring $\that$ is close to $\vtheta_\alpha$ and ensuring that the missingness mechanism has a small expected missingness rate.


In Appendix~\ref{sxn:optimal_adversarial_scm_selection}, a characterization of the optimal adversarial SCM is given for an arbitrarily parameterized family of SCMs, assuming that the adversarial DAG is a subgraph of the true DAG. In the special case of linear Gaussian SCMs, this leads to a closed form solution for the optimal adversarial SCM. This result is used to select adversarial SCMs in our experimental evaluations.

\section{Adversarial Missingness via Rejection Sampling}
\label{sxn:generalized_rejection_sampling}

Our first result establishes a general procedure for guiding modelers that optimize~\eqref{eqn:modeler-objective} to produce $\that = \vtheta_\alpha$, when the adversary has access to a $\beta$-indistinguishable adversarial SCM. The approach uses rejection sampling, so the bound on the missingness rate is implicitly determined by the relationship between $\P_{\rvx;\vtheta_\alpha}$ and $\P_{\rvx;\vtheta_p}$. A general setup for this rejection sampling approach is given in Appendix~\ref{sxn:proofs_generalized_rejection_sampling} that is appropriate for removing multiple edges. Here we consider a local variant that is appropriate for removing one or a small number of edges, and that has a more favorable missingness rate.

Let $\mathcal{V} \subseteq \{1, \ldots, d\}$ denote a subset of the variables, and $\overline{\mathcal{V}}$ denote the complement. Localized generalized rejection sampling on the variables $\mathcal{V}$ is a missingness mechanism that masks only variables in $\mathcal{V}$, using probabilities depending only on the value of $\rvx_{\mathcal{V}}$, given by

\begin{equation}\label{eqn:localized_generalized_rejection_sampling}
\P_{\rvr|\rvx}(\vr|\vx) = \begin{cases}
\frac{1}{2^{|\mathcal{V}|}-1} \frac{\Lambda(\vx_{\mathcal{V}})}{\Lambda} & \text{ if } \vr_{\overline{\mathcal{V}}}=1 \text{ and } \vr_{\mathcal{V}} \neq 0 \\
1 - \frac{\Lambda(\vx_{\mathcal{V}})}{\Lambda} & \text{ if } \vr_{\overline{\mathcal{V}}}=1 \text{ and } \vr_{\mathcal{V}} = 0 \\
0 & \text{ otherwise}
\end{cases}.
\end{equation}
Here, $\Lambda(\vx_{\mathcal{V}}) = \frac{\p_{\rvx_{\mathcal{V}};\vtheta_\alpha}(\vx_{\mathcal{V}})} {\p_{\rvx_{\mathcal{V}};\vtheta_p}(\vx_{\mathcal{V}})} $ is the ratio of the adversarial distribution to the true distribution, and $\Lambda = \max_{\vx_{\mathcal{V}}} \Lambda(\vx_{\mathcal{V}})$ is the maximum value of that ratio. Note that the observation patterns that select all variables in $\overline{\mathcal{V}}$ and at least one variable in $\mathcal{V}$ are equiprobable. Because this approach only drops variables in $\mathcal{V}$, the missingness rate is at most $\tfrac{|\mathcal{V}|}{d}$. Lemma~\ref{lem:localized_generalized_rejection_sampling_missingness_bound} in the Appendix establishes a tighter bound on the missingness rate that depends on the ratio $\Lambda$. 

When the conditional distributions of the variables in $\overline{\mathcal{V}}$ given the variables in $\mathcal{V}$ is identical in the adversarial and true SCMs, localized generalized rejection sampling ensures that the partially observed features from $\P_{\rvx;\vtheta_p}$ look as though they were sampled from the adversarial distribution.

\begin{restatable}[Localized Rejection Sampling]{lemma}{localizedrejectionsampling}
\label{lem:localized_generalized_rejection_sampling}
Let $\mathcal{V} \subset \{1, \ldots, d\}$ be a subset of the variables. If it is the case that the adversarial distribution preserves the dependence of $\overline{V}$ on $\mathcal{V}$, that is,
\[
\P_{\rvx_{\overline{\mathcal{V}}} \mid \rvx_{\mathcal{V}}}( \cdot \mid  \cdot\,; \vtheta_\alpha) = \P_{\rvx_{\overline{\mathcal{V}}} \mid \rvx_{\mathcal{V}}}(\cdot \mid \cdot\, ; \vtheta_p),
\] and the adversary uses the missingness mechanism defined  in~\eqref{eqn:localized_generalized_rejection_sampling}, then
\begin{equation}
\label{eqn:matched_conditional_distribution}
\P_{\rvx_o\mid \rvr}(\cdot\,\mid \vr; \vtheta_p) = \P_{\rvx_o}(\cdot\,; \vtheta_\alpha)
\end{equation}
for all $\vr$ such that $\P_{\rvr}(\vr) \neq 0$.
\end{restatable}

Proof is given in Appendix \ref{sxn:proofs_generalized_rejection_sampling}. This result implies that when the matching condition~\eqref{eqn:matched_conditional_distribution} holds, the adversary can attain their goal of causing $\vtheta_\alpha$ to be a global maximizer of the modeler's objective.

\begin{restatable}{corollary}{attacksuccesslocalized}\label{cor:attack_success_localized_generalized_rejection_sampling}
If it is the case that the adversarial distribution satisfies
\[
\P_{\rvx_{\overline{\mathcal{V}}} \mid \rvx_{\mathcal{V}}}( \cdot \mid  \cdot\,; \vtheta_\alpha) = \P_{\rvx_{\overline{\mathcal{V}}} \mid \rvx_{\mathcal{V}}}(\cdot \mid \cdot\, ; \vtheta_p)
\]
and the adversary uses localized rejection sampling (\eqref{eqn:localized_generalized_rejection_sampling}), then $\vtheta_\alpha$ is a global maximizer of the objective of the modeler (\eqref{eqn:modeler-objective}).
\end{restatable}

This result implies, in particular, that if the adversary's goal is to delete a subset of the incoming edges to a node $s$ and the adversarial SCM is constructed such that the parents of $s$ in $\mathcal{G}_\alpha$ are a subset of the parents of $S$ in $\mathcal{G}$, and all other causal relationships in the SCMs are identical, then the adversarial distribution is a global maximizer of the modeler's objective when localized rejection sampling is used with $\mathcal{V} = \{s\} \cup \pa_s$. This fact is established as Corollary~\ref{cor:local_edge_deletion} in Appendix~\ref{sxn:proofs_generalized_rejection_sampling}.


\section{Learned Adversarial Missingness Mechanism (LAMM)}
\label{sxn:neural_approach}

The rejection sampling approaches can be applied to finite training data sets, but offer little control of the missingness rate and their optimality guarantees hold when the modeler can evaluate the expectations involved in its objective. It is attractive to consider approaches that tailor the adversarial missingness mechanism specifically to the finite training data set at hand, and that explicitly encourage the missingness rate to be low.

With finite data, the expectations in~\eqref{eqn:modeler-objective} must be replaced with empirical averages. Moreover, even for SCMs parameterized with exponential family distributions, the objective is non-concave due to the presence of missing data, which leads practitioners to use the Expectation Maximization (EM) algorithm to learn the parameters~(\cite{missdag,MissGLasso2010}). In this setting, the adversary's goal is to select a missingness distribution such that EM converges to the adversarial parameter $\ta$.

To that end, we propose to parameterize the missingness distribution with a neural network. Let $\midx$ denote the variables the adversary chooses for local masking and $y(\vx_\midx,\vphi)=\text{softmax}((f^{L}\circ \dots \circ f^{1})(\vx_\midx))$ be an $L$-hidden layer neural network with $2^{|\midx|}$ output units; here
$\vphi$ are the parameters of the network. Each output unit returns the probability of one of the observation patterns $\vr_{\mathcal{V}}$. Let $\delta:\{0,1\}^{|\midx|}\rightarrow \{0,1,\dots,2^{|\midx|}-1\}$ denote the function that maps the observed mask pattern to the corresponding output neuron. Then the missingness distribution is parameterized as follows:
\begin{equation}
\P_{\rvr|\rvx}(\vr|\vx,\vphi)= 
\begin{cases}
    y(\vx_\midx,\vphi)_{\delta(r_{\midx})},&\text{if } \vr_{\overline{\midx}}=1\\
0,              &\text{otherwise}
\end{cases}
\end{equation}

Our goal is to optimize the adversary's objective, \eqref{eqn:adversarial-objective2} by choosing $\vphi$ appropriately. Recall that the modeler, given partially observed data sampled according to the adversary's missingness mechanism, chooses the parameter $\that$ in~\eqref{eqn:adversarial-objective2} by solving its own objective~\eqref{eqn:modeler-objective}. In order to learn an optimal $\vphi$ to parametrize the adversary's missingness mechanism, we model the dependence of $\that$ on $\vphi$ in a differentiable manner.

The EM algorithm is the canonical approach to find a (approximately) minimizing $\that$ for~\eqref{eqn:modeler-objective} given a single sampled realization of the missingness mechanism, but because of the sampling process, its output $\that$ is not differentiable with respect to $\vphi$. Instead of sampling from the missingness mechanism, we take the expectation with respect to it; this results in a differentiable objective. We call this formulation the Weighted EM (WEM) algorithm. Due to space constraints, the details of the expectation and maximization steps of the WEM algorithm are given in Appendix~\ref{app:wem}.

Given that the modeler's procedure for optimizing its objective to learn $\that$ is captured by the WEM algorithm, the adversary's goal is to make WEM converge to $\ta$ from an arbitrary starting point $\vtheta^{0}$. Denote the corresponding output of the WEM algorithm (Algorithm~\ref{alg:wem} in Appendix~\ref{app:wem}) by $\text{WEM}(\vphi,\vtheta^{0},\epsilon,\data)$. The adversary solves the following problem using gradient-based optimization:
\begin{align}
\label{eq:wem2}
		\min_{\vphi } & \quad \text{D}_{\text{KL}}(\P_{\rvx;\vtheta_\alpha} \| \P_{\rvx;\ttil}) + \frac{\lambda}{N}\sum_{i=1}^N \E_{\rvr|\rvx;\vphi} \left[\frac{|\{j\,|\,\rvr_j=0\}|}{d}\,\Big\lvert\, \rvx=\vx^{(i)}\right] \\
		&\kern-2em \text{where } \ttil =\text{WEM}(\vphi,\vtheta^{0},\epsilon,\data).\notag
\end{align}
This formulation accounts for the adversary's desire to bound the expected missingness rate in~\eqref{eqn:adversarial-objective2}. This objective is exactly~\eqref{eqn:adversarial-objective2}, except optimization with respect to the missingness distribution $\P_{\rvr|\rvx}$ has been replaced with optimizing with respect to $\phi$, which parameterizes the missingness distribution. In general, solving this optimization problem requires the WEM algorithm to be started from scratch in each training epoch using the updated weights, but for exponential family distributions, the maximization step of WEM admits a simple form. Details are given in Appendix~\ref{app:wem}. 

In practice, the modeler's initialization scheme is unknown, and the adversary could overfit to a particular initialization when solving~\eqref{eq:wem2}. To mitigate this, $\vphi$ is selected to guide multiple random initializations to $\ta$. The proposed method of learning a missingness distribution is described in Algorithm~\ref{alg:nn}. The LAMM formulation is flexible and, when more is known about the subroutine the modeler is using to learn $\that$, one can replace WEM with an appropriate differentiable subroutine. 

\begin{algorithm}[tb]
	\caption{LAMM Algorithm. Learns the adversarial missingness mechanism by directing weighted EM to converge on a desired parameter. The subroutine WEM is described in Algorithm~\ref{alg:wem} of Appendix~\ref{app:wem}. $\ell (\ttil_k, \ta,\vphi,\lambda)$ denotes the objective function given in \eqref{eq:wem2}}.
	\label{alg:nn}
	
	\begin{algorithmic}
		\STATE {\bfseries Input: $\vtheta_\alpha, \epsilon, K, \data$, $\lambda$}
		\STATE $\bm{\phi} \leftarrow \text{Initialize}$
		\STATE $\vtheta^{0}_k\leftarrow \text{Initialize } \ k=1,\dots,K$ 
		\WHILE{$\bm{\phi}$ not converged}
		\STATE  \COMMENT{Using $K$ starting points for robustness}
		\FOR{\texttt{$k=1,\dots,K$}}
		\STATE $\tilde{\vtheta}_k \leftarrow \text{WEM}(\vphi, \vtheta^{0}_k, \epsilon, \data)$ \COMMENT{WEM Algorithm}
		\ENDFOR
		\STATE $\bm{\phi} \leftarrow \bm{\phi}- \frac{\eta }{K}\sum_{k=1}^K\nabla_{\bm{\phi}} (\ell (\ttil_k, \ta,\vphi,\lambda))$
		\ENDWHILE
		\STATE {\bfseries Output: $\bm{\phi}$} 
	\end{algorithmic}
\end{algorithm}

\section{Experiments}\label{sec:exp}

The experimental setup has three components corresponding to parts of the adversarial missingness threat model shown in Figure~\ref{fig:overall}: the underlying true SCM ($\tr$); the causal discovery algorithm employed by the modeler; and the adversary's choice of the adversarial DAG, adversarial SCM, and missingness mechanism ($\ga,\ta,\prx$).

\begin{table}[b]
	\centering
	\caption{Overview of the experiments conducted.}
	\resizebox{.5\columnwidth}{!}{%
\begin{tabular}{|l|r|r|r|r|}
\toprule
\multicolumn{1}{|c|}{\multirow{2}[4]{*}{Name}} & \multicolumn{2}{c|}{$\mathcal{G}_{p}$} & \multicolumn{1}{c|}{\multirow{2}[4]{*}{N}} & \multicolumn{1}{c|}{\multirow{2}[4]{*}{Target Edge}} \\
\cmidrule{2-3}      & \multicolumn{1}{c|}{d} & \multicolumn{1}{c|}{\# Edges} &       &  \\
\midrule
Gaussian SCM, I & 3     & 2     & 1k    & $1\rightarrow 2$ \\
\midrule
Gaussian SCM, II & 6     & 5     & 50k   & $2\rightarrow 3$ \\
\midrule
Sachs Dataset & 11    & 17    & 853   & ``plc'' $\rightarrow$ ``pip2'' \\
\bottomrule
\end{tabular}%

	}
	\label{tab:experiment_stats}%
\end{table}%

\textbf{The true SCM.} For the true SCMs we have used linear Gaussian SCMS with equal noise variance, as they are a popular choice (e.g. \cite{missdag, notears,golem}). Two of the experiments use simulated data, and one utilizes the commonly used Sachs dataset~\cite{sachs}. In each experiment, a single edge was targeted for deletion via adversarial missingness. Salient characteristics of the experiments are provided in Table~\ref{tab:experiment_stats}. 

\textbf{Modeler's Causal Structure Learning Algorithm.} For the modeler's causal structure learning algorithms we have employed methods developed for learning in the presence of missing data, namely the MissDAG \cite{missdag} (denoted as MissDAG (NT)), a score based method, and MissPC~\cite{misspc}, a constraint-based method. We have also compared with approaches that use mean imputation followed by structure learning algorithms that require fully observed data; namely, we have utilized mean imputation followed by the NOTEARS~\cite{notears} and PC algorithms. MissDAG is sensitive to the initialization of the parameters,  so we initialized the covariance matrices using five different schemes: Empirical Diagonal (``Emp. Diag.''), identity matrix (``Ident.''), the ground truth covariance matrix (``True''), a scaled random covariance matrix (``Random(*)'') and a scaled inverse Wishart random matrix (``IW(*)''). For details refer to Appendix \ref{app:init}.

\textbf{Adversary's Choices}. In each experiment, we describe how $\ga$ and the adversarial SCM are selected. The adversarial missingness distribution $\prx$ (denoted as MNAR in the tables) is either a variation of generalized local rejection sampling (\eqref{eqn:localized_generalized_rejection_sampling}) or selected via the \namenn algorithm (Algorithm \ref{alg:nn}). To test the relative advantages of our adversarial missingness mechanisms, we also employ missingness distributions that drop an equal amount of data completely at random. These distributions are denoted MCAR in the tables. To match the amount of missing data and the marginal distribution over the observation masks, the MCAR distributions are taken to be the marginal distribution of $\rvr$ given the MNAR missingness distribution $\P_{\rvr \mid \rvx}$ , i.e., $\P_{\rvr}(\vr) = \E_{\rvx; \vtheta_p}[ \P_{\rvr \mid \rvx}(\vr)]$. 


\textbf{Performance Metrics} To measure the performance of the adversarial missingness attacks, we first sample data from the true SCM and generate missing data masks ${\vr^{(i)} \mid \vx^{(i)}\sim \prx(\cdot\,;\,\vx^{(i)})}$ according to the relevant missingness mechanism, and generate masked data sets $\{\hat{\vx}^{(i)}\}_{i=1}^N$ where $\hat{\vx}^{(i)}$ has observed values in $\vr^{(i)}$ and NaNs to denote the missing entries. Given the partially observed data, we employ the relevant causal structure learning algorithm to estimate an SCM with corresponding DAG $\mathcal{\hat{G}}$. We report the Hamming distance (HD), the number of edge differences, between $\mathcal{\hat{G}}$ and $\mathcal{G}_p$. If $\mathcal{\hat{G}}$ is a partial DAG, for each edge in the true graph, partial DAG has to contain the corresponding undirected edge. The adversarial attack is deemed successful if the edge targeted for deletion is not present in $\mathcal{\hat{G}}$. To account for the randomness in the missing data masks, this process is repeated multiple times and the average success rate and HD are reported.

All experiments involving a neural network employ a 2-hidden layer network with ReLu activation and 100 units per layer; see Appendix \ref{app:nn} for implementation details. 

\paragraph{Simulation Experiments.}
In our two simulation experiments, we have used a Gaussian SCM, $\rvx = \mB^T \rvx + \vn$,
where $\vn$ comprises independent Gaussian noise distributed as $\mathcal{N}(0, \mI)$.

In both experiments the adversarial goal is to remove a single edge. Let $(p,c)$ denote the parent and the child nodes corresponding to the removed edge. The removed edge $\wa_{i,j}=0$ is set to zero unless otherwise stated. Following Theorem~\ref{thm:optimal_adversarial_linear_gaussian_SCM}, the parameters $\mB^\alpha$ are kept the same as those of the true SCM except at the removed edge i.e. $\wa_{i,j}=\mB_{i,j}$ for all $(i,j)\neq (p,c)$ and $\sa_j=1$ for all $j\neq c$. We take $\sa_c=\sigma$, which is not optimal in terms of the KL-divergence between the adversarial SCM and the true SCM, but is consistent with the modeler's assumptions of equal variance.

\paragraph{Gaussian SCM, I.}
We designed this experiment as a feasibility check of the LAMM approach with the local masking. The graph has three nodes, and nodes 2 and 3 have incoming edges from node 1 (See Appendix \ref{app:scm1}). These edges have magnitudes $0.8$ and $0.9$, both of which are above the minimum threshold set in~\cite{notears} to eliminate spurious edges. This $\mB$ results in a Pearson correlation of roughly 0.6 between $\rvx_1$ and $\rvx_2$, which makes localized generalized rejection sampling feasible (as the maximum of the density ratio is kept reasonable) but non-trivial. 

The adversarial goal is to remove the edge from 1 to 2 i.e. $(p,c)=(1,2)$ and the adversary employs LAMM with $\lambda=1e-2$. Following Lemma~\ref{lem:localized_generalized_rejection_sampling}, only nodes 1 and 2 are masked, so $\midx=\{1,2\}$. See Table \ref{tab:scm1}) for the experimental results. The trained missingness mechanism masks 40\% of the entries in the target two columns on average; the overall percentage of the training data masked is 27\%. LAMM ensures that the modeler converges to $\ga$ in all modeler configurations including when the modeler uses PC-based causal structure learning algorithms. These algorithms are very different from the EM-based algorithms that LAMM was designed to target. This experiment also suggests that the missingness mask must be selected on a per observation basis, as the MCAR missingness distribution never achieves the adversarial goal.

\paragraph{Gaussian SCM, II.}
This experiment is designed to gauge the effectiveness of the localized rejection attack in Lemma~\ref{lem:localized_generalized_rejection_sampling}. We slighlty modified the equation for practical purposes (as described in Appendix~\ref{app:scm2}).

The true SCM is a 6 node graph with 5 edges, and the adversary aims to remove a single edge. The edges weights are selected randomly between $[0.5,2]\cup[-2,-0.5]$ as in~\cite{notears}, except for the edge between nodes 2 and 3.  To ensure the acceptance probability is at a reasonable level, we selected a low weight for the target edge: $\mB_{2,3}=0.4$. The edge weights, $\mB$, are given in Appendix~\ref{app:scm2}. 

The weight of the target edge is set to 0.25 in $\wa$, i.e. $\wa_{2,3}=0.25$, to increase the acceptance probability and because the NT algorithm uses a weight threshold at 0.3, as suggested by the authors~\cite{notears}.

The adversary utilizes the local rejection sampling algorithm, and following Lemma~\ref{lem:localized_generalized_rejection_sampling}, masks only $\midx=\{2,3\}$. In 
this experiment, the accepted samples are fixed to be fully observed. This results in $70\%$ of the rows of the two target columns missing entirely. The results, given in Table 3, show that the adversary achieves its goal when the modeler uses the missDAG algorithm, but not when the modeler uses the PC algorithm. This perhaps because the PC algorithm's significance threshold is lower than that of the NT algorithm. Although $\wa_{2,3}$ could be lowered, this would increas the maximum of the density ratios and result in unacceptable levels of missing data. 

\begin{table}[htbp]
	\centering
	\caption{Results for Gaussian SCM, I. Average performances are reported using 50 different mask samples. \namenn always removes the target edge while MCAR never does. \namenn does not introduce any extraneous edges as HD($\hat{\mathcal{G}},\mathcal{G}_{p}$) is always one.  }
	\resizebox{0.6\columnwidth}{!}{%
\begin{tabular}{|l|c|l|l|l|l|}
\toprule
\multicolumn{1}{|c|}{\multirow{2}[4]{*}{Modeler}} & \multirow{2}[4]{*}{Initial.} & \cellcolor[rgb]{ 1,  .753,  0}MNAR (\namenn) & \cellcolor[rgb]{ 0,  .69,  .941}MCAR & \cellcolor[rgb]{ 1,  .753,  0}MNAR (\namenn) & \cellcolor[rgb]{ 0,  .69,  .941}MCAR \\
\cmidrule{3-6}      &       & HD($\hat{\mathcal{G}},\mathcal{G}_{p}$) & HD($\hat{\mathcal{G}},\mathcal{G}_{p}$) & \multicolumn{1}{p{5em}|}{ Success} & \multicolumn{1}{p{5em}|}{ Success} \\
\midrule
\multicolumn{1}{|c|}{\multirow{5}[10]{*}{MissDAG}} & \multicolumn{1}{l|}{Emp. Diag.} & 1     & \textbf{0} & \textbf{1} & 0 \\
\cmidrule{2-6}      & \multicolumn{1}{l|}{IW(*)} & 1     & \textbf{0} & \textbf{1} & 0 \\
\cmidrule{2-6}      & \multicolumn{1}{l|}{Ident.} & 1     & \textbf{0} & \textbf{1} & 0 \\
\cmidrule{2-6}      & \multicolumn{1}{l|}{Random(*)} & 1     & \textbf{0} & \textbf{1} & 0 \\
\cmidrule{2-6}      & \multicolumn{1}{l|}{True} & 1     & \textbf{0} & \textbf{1} & 0 \\
\midrule
MissPC & -     & 1     & \textbf{0.02} & \textbf{1} & 0 \\
\midrule
Mean + NT & -     & \textbf{1} & \textbf{1} & \textbf{1} & 0 \\
\midrule
Mean + PC & -     & 1     & \textbf{0.22} & \textbf{1} & 0 \\
\bottomrule
\end{tabular}%

	}
	\label{tab:scm1}%
\end{table}%

\begin{table}[htbp]
	\centering
	\caption{Results of Gaussian SCM, II. Average performances are reported using 20 different mask samples. RS-based missingness attacks are more successful than their MCAR counterparts, but due to the high amount of missing data, even MCAR missingness can lead to a $100\%$ success rate for certain missDAG initializations. }
	\resizebox{0.6\columnwidth}{!}{%
\begin{tabular}{|l|c|r|r|r|r|}
\toprule
\multicolumn{1}{|c|}{\multirow{2}[4]{*}{Modeler}} & \multirow{2}[4]{*}{Initial.} & \multicolumn{1}{l|}{\cellcolor[rgb]{ 1,  .753,  0}MNAR (RS)} & \multicolumn{1}{c|}{\cellcolor[rgb]{ 0,  .69,  .941}MCAR} & \multicolumn{1}{l|}{\cellcolor[rgb]{ 1,  .753,  0}MNAR (RS)} & \multicolumn{1}{c|}{\cellcolor[rgb]{ 0,  .69,  .941}MCAR} \\
\cmidrule{3-6}      &       & \multicolumn{1}{l|}{HD($\hat{\mathcal{G}},\mathcal{G}_{p}$)} & \multicolumn{1}{l|}{HD($\hat{\mathcal{G}},\mathcal{G}_{p}$)} & \multicolumn{1}{p{5em}|}{ Success} & \multicolumn{1}{p{5em}|}{ Success} \\
\midrule
\multicolumn{1}{|c|}{\multirow{5}[10]{*}{MissDAG}} & \multicolumn{1}{l|}{Emp. Diag.} & \textbf{2} & \textbf{2} & \textbf{1} & \textbf{1} \\
\cmidrule{2-6}      & \multicolumn{1}{l|}{IW(*)} & 2     & \textbf{1.45} & \textbf{1} & 0.4 \\
\cmidrule{2-6}      & \multicolumn{1}{l|}{Ident.} & 2     & \textbf{1.85} & \textbf{1} & 0.85 \\
\cmidrule{2-6}      & \multicolumn{1}{l|}{Random(*)} & 2     & \textbf{1.6} & \textbf{1} & 0.4 \\
\cmidrule{2-6}      & \multicolumn{1}{r|}{TRUE} & 2     & \textbf{1} & \textbf{1} & 0 \\
\midrule
MissPC & -     & \textbf{0.05} & \textbf{0.05} & \textbf{0.05} & 0 \\
\midrule
Mean + NT & -     & \textbf{5} & \textbf{5} & \textbf{0.6} & 0 \\
\midrule
Mean + PC & -     & 6.7   & \textbf{5.5} & \textbf{0} & \textbf{0} \\
\bottomrule
\end{tabular}%
	}
	\label{tab:addlabel}%
\end{table}%

\paragraph{Sachs Dataset.}
We used observational data from~\cite{sachs} to test our methods in a challenging setting. The data set contains only $N=853$ samples from a system with 11 different variables, and the ground truth SCM has 17 edges. This data set has posed a challenge to causal discovery algorithms even in the fully observed case (\cite{golem},\cite{zhuCausalDiscoveryReinforcement2020}). For this reason, our adversarial goal is to remove a correct edge from the DAG estimated from the fully observed data. 

\begin{figure}[ht]
	\centering
	\includegraphics[width=0.5\columnwidth]{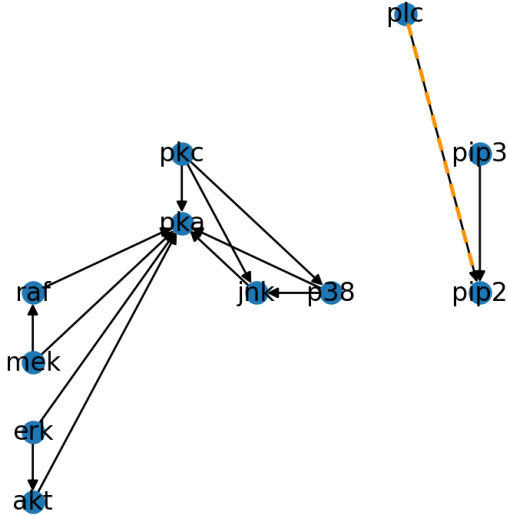}
	\caption{NT estimated graph in the Sachs dataset and two connected components. Dashed orange edge denotes the adversarial target.}
	\label{fig:nt}
\end{figure}
The NT algorithm estimated a DAG with 12 HD to the ground truth DAG (Figure \ref{fig:nt}) and managed to capture the two connected components present in the true DAG. Our adversarial goal is to remove the correctly estimated edge from ``plc'' to ``pip2''. In our formulation the ``true graph'' $\gp$ is the one estimated by the NT algorithm from the fully observed data.

Since this is a real dataset, the true SCM parameters $\tr$ are unknown, so we used the empirical covariance matrix while selecting the adversarial parameter in a heuristic way\footnote{The mean vector is set to zero after subtracting the average values from each column, following the assumptions in NT.}. Removing the edge between ``plc'' to ``pip2'' makes ``plc'' an isolated node, so we set the covariance terms from ``plc'' to ``pip2'' and ``plc'' to ``pip3'' zero. This approach is heuristic, but we observed that the covariance matrix corresponding to the NT estimated $\mB$ did not match the empirical covariance matrix accurately. It suggests Gaussian SCM model might be inaccurate and using $\mB$ directly may lead to unwanted changes in the distribution.

The adversary uses \namenn with $\lambda=0$ and following Lemma~\ref{lem:localized_generalized_rejection_sampling} masks only $\midx=\{\text{``plc'',``pip2'',``pip3''}\}$. The missingness distribution learned masks $51.0\%$ of the three masked variables, which corresponds to a missingness rate of $13.9\%$ over all the variables (See loss function Appendix Figure \ref{fig:loss}). 

The results are displayed in Table \ref{tab:sachs}. For missDAG and NT after mean imputation, \namenn has a higher success rate with relatively less unintentional edges added. \namenn has its lowest success rate ($70\%$) against missDAG with the random initializations. We also observed for missPC, even MCAR missingness has a $100\%$ success rate. This suggests that the PC algorithm does not converge to the graph that NT estimates from the fully observed data.

\begin{table}[htbp]
	\centering
	\caption{Results for the Sachs dataset. Average results are reported using 20 different mask samples. \namenn is consistently more successful than MCAR and has a lower distance to the true graph. Random initializations cause \namenn to reach its highest distance from the reference graph and its lowest success rate.}
	\resizebox{0.6\columnwidth}{!}{%
\begin{tabular}{|l|c|r|r|r|r|}
\toprule
\multicolumn{1}{|c|}{\multirow{2}[3]{*}{Modeler}} & \multirow{2}[3]{*}{Initial.} & \multicolumn{1}{l|}{\cellcolor[rgb]{ 1,  .753,  0}MNAR (\namenn)} & \multicolumn{1}{l|}{\cellcolor[rgb]{ 0,  .69,  .941}MCAR} & \multicolumn{1}{l|}{\cellcolor[rgb]{ 1,  .753,  0}MNAR (\namenn)} & \multicolumn{1}{l}{\cellcolor[rgb]{ 0,  .69,  .941}MCAR} \\
\cmidrule{3-6}      &       & \multicolumn{1}{l|}{HD($\hat{\mathcal{G}},\mathcal{G}_{p}$)} & \multicolumn{1}{l|}{HD($\hat{\mathcal{G}},\mathcal{G}_{p}$)} & \multicolumn{1}{p{5em}|}{ Success} & \multicolumn{1}{p{5em}|}{ Success} \\
\multicolumn{1}{|c|}{\multirow{5}[9]{*}{MissDAG}} & \multicolumn{1}{l|}{Emp. Diag.} & \textbf{2.95} & 4.85  & \textbf{0.95} & 0.4 \\
\cmidrule{2-6}      & \multicolumn{1}{l|}{IW(*)} & \textbf{4.25} & 6.75  & \textbf{0.7} & 0.5 \\
\cmidrule{2-6}      & \multicolumn{1}{l|}{Ident.} & \textbf{3.3} & 4.85  & \textbf{0.95} & 0.35 \\
\cmidrule{2-6}      & \multicolumn{1}{l|}{Random(*)} & \textbf{4.05} & 5.8   & \textbf{0.7} & 0.15 \\
\cmidrule{2-6}      & \multicolumn{1}{r|}{TRUE} & \textbf{2.95} & 4.55  & \textbf{0.95} & 0.15 \\
\midrule
MissPC & -     & 10.1  & \textbf{9.45} & \textbf{1} & \textbf{1} \\
\midrule
Mean + NT & -     & \textbf{2.4} & 3.4   & \textbf{1} & 0.45 \\
\midrule
Mean + PC & -     & \textbf{9.05} & 9.6   & \textbf{1} & 0.65 \\
\bottomrule
\end{tabular}%
	}
	\label{tab:sachs}%
\end{table}%

\section{Conclusion}

This work introduced the adversarial missingness model for influencing the learning of structural causal models from data. This adversarial model is appropriate in settings where attempts by the adversary to manipulate the values of the data can be detected, and the ubiquity of benignly missing data supports the use of adversarial missingness as a vector of attack. 

Generalized rejection sampling schemes were introduced and proven to achieve many of the desiderata of adversarial missingness, thereby establishing a strong proof of concept of the threat model. As a practical methodology for AM with finite training data sets, we provided a heuristic for learning adversarial missingness mechanisms, and demonstrated its performance using data drawn from synthetic and real SCMs.

Many aspects of the AM threat model remain to be explored, e.g.: (1) can one design algorithms that provably achieve the desiderata of AM in the finite data setting, (2) can one quantify the trade-offs between the desiderata of the adversary (e.g. the missingness rate and the attack success or the missing rate and the $\beta$-indistinguishability), and (3) how modelers defend against AM attacks? We expect that meaningful answers to these questions depend on the functional form of the SCMs under consideration, and are currently investigating these issues in the context of linear Gaussian SCMs.

\bibliographystyle{abbrv}
\bibliography{references}

\clearpage
\appendix
\section{Characterization and Selection of Optimal Adversarial SCMs}
\label{sxn:optimal_adversarial_scm_selection}

Rejection sampling provides a methodology for conducting adversarial missingness attacks on arbitrarily parameterized SCMs, \emph{if} the target adversarial SCM is known \emph{a priori} to satisfy $\beta$-indistinguishability and to be Markovian with respect to the adversarial DAG $\mathcal{G}_\alpha$. Similarly, the LAMM heuristic provides a sample efficient methodology under the same assumptions.

However, given the true SCM in an arbitary parameterized family of SCMs, an arbitrary $\beta$, and an arbitary adversarial DAG $\mathcal{G}_\alpha$, it is nontrivial to even determine whether a $\beta$-indistinguishable SCM that is Markovian with respect to $\mathcal{G}_\alpha$ exists, much more to find one. In this section, we provide a general characterization of optimal adversarial SCMs when $\mathcal{G}_\alpha$ is a subgraph of $\mathcal{G}_p$, i.e. a characterization of the SCMs closest to $\P_{\rvx; \vtheta_p}$ in KL-divergence while also being Markov with respect to $\mathcal{G}_\alpha$. In the case of linear Gaussian SCMs, these characterizations provide a practical approach to finding an optimal SCM.

The key observation to obtaining our characterization is that the KL-divergence between two distributions that are Markov with respect to the same DAG satisfies a convenient factorization.

\begin{lemma}
\label{lem:factorize_KL}
If $\P_{\rvx; \vtheta_1}$ and $\P_{\rvx; \vtheta_2}$ are SCMs that are Markovian with respect to the DAG $\mathcal{G}$, then 
\begin{equation*}
 \operatorname{D}_{\mathrm{KL}}\left( \P_{\rvx;\vtheta_1} \,\|\, \P_{\rvx; \vtheta_2} \right) =
 \sum_{j=1}^d \E_{\rvx_{\pa_j};\vtheta_1} \left[ \operatorname{D}_{\mathrm{KL}}\left( \P_{\rvx_j\mid \rvx_{\pa_j}; \vtheta_1} \,\|\, \P_{\rvx_j \mid \rvx_{\pa_j}; \vtheta_2} \right) \right].
\end{equation*}
\end{lemma}

\begin{proof}
    Write the causal factorizations of the pdfs of the two distributions with respect to $\mathcal{G}$ as
$$
\p(\vx; \vtheta_1) = \prod_{j=1}^d p(\vx_j\,|\,\vx_{\pa_j}; \vtheta_1) \quad \text{and} \quad \p(\vx; \vtheta_2) = \prod_{j=1}^d p(\vx_j\,|\,\vx_{\pa_j}; \vtheta_2),
$$
and observe that, as a consequence of these factorizations, the KL-divergence satisfies
\begin{align*}
    \operatorname{D}_{\mathrm{KL}}\left( \P_{\rvx;\vtheta_1} \,\|\, \P_{\rvx; \vtheta_2} \right) & = 
    \E_{\rvx; \vtheta_1} \left[ \ln\left( \frac{\p(\vx; \vtheta_1)}{\p(\vx; \vtheta_2)} \right) \right] \\
   &  \kern-5em = \E_{\rvx; \vtheta_1} \left[ \sum_{j=1}^d \ln\left( \frac{\p(\vx_j \mid \vx_{\pa_j}; \vtheta_1)}{\p(\vx_j \mid \vx_{\pa_j} ; \vtheta_2)} \right) \right] \\
    & \kern-5em = \sum_{j=1}^d \E_{\rvx_{\pa_j \cup \{j\}}; \vtheta_1} \left[ \ln\left( \frac{\p(\vx_j \mid \vx_{\pa_j}; \vtheta_1)}{\p(\vx_j \mid \vx_{\pa_j} ; \vtheta_2)} \right) \right] \\
    & \kern-5em = \sum_{j=1}^d \E_{\rvx_{\pa_j}; \vtheta_1} \left[ \E_{\rvx_j \mid \rvx_{\pa_j}; \vtheta_1} \left[  \ln \left( \frac{\p(\vx_j \mid \vz; \vtheta_1)}{\p(\vx_j \mid \vz ; \vtheta_2)} \right) \,\Big\lvert\, \rvx_{\pa_j} = \vz \right] \right] \\
    & \kern-5em = \sum_{j=1}^d \E_{\rvx_{\pa_j}; \vtheta_1} \left[ \operatorname{D}_{\mathrm{KL}}( \P_{\rvx_j\mid \rvx_{\pa_j}; \vtheta_1} \,\|\, \P_{\rvx_j \mid \rvx_{\pa_j}; \vtheta_2} )\right].
\end{align*}
The third equation holds because the structural equation for $\rvx_j$ involves only the variables $\rvx_j$ and $\rvx_{\pa_j}$, and the following equation uses the tower property of conditional expectation.
\end{proof}

The preceding result suggests that when the adversarial DAG is a subgraph of $\mathcal{G}_p$, then the adversarial SCM that minimizes the KL-divergence from the true SCM will change a minimal number of structural equations: in particular, when the parents of $\rvx_j$ are the same in both graphs, using the same structural equation for $\rvx_j$ ensures that
\[
\E_{\rvx_{\pa_j};\vtheta_1} \left[ \operatorname{D}_{\mathrm{KL}}\left( \P_{\rvx_j\mid \rvx_{\pa_j}; \vtheta_1} \,\|\, \P_{\rvx_j \mid \rvx_{\pa_j}; \vtheta_2} \right) \right]= 0.
\]
This claim is a specific instance of a more general result characterizing the adversarial SCM that minimizes the KL-divergence from the true SCM.

\begin{corollary}
    \label{cor:adversarial_optimality_condition}
	Suppose $\mathcal{G}_\alpha$ is a subgraph of $\mathcal{G}_p$. If $\P_{\rvx; \vtheta_\alpha}$ is Markov relative to $\mathcal{G}_\alpha$ and satisfies
 \begin{equation*}
 \E_{\rvx_{\pa_j};\vtheta_p} \left[ \operatorname{D}_{\mathrm{KL}}( \P_{\rvx_j \mid \rvx_{\pa_j};\vtheta_p}  \,\|\, \P_{\rvx_j \mid \rvx_{\pa_j}; \vtheta_\alpha}) \right] \leq
  \E_{\rvx_{\pa_j};\vtheta_p} \left[ \operatorname{D}_{\mathrm{KL}}( \P_{\rvx_j \mid \rvx_{\pa_j};\vtheta_p}  \,\|\, \P_{\rvx_j \mid \rvx_{\pa_j}; \vtheta}) \right]
 \end{equation*}
 for all $\P_{\rvx; \vtheta}$ that are Markov relative to $\mathcal{G}_\alpha$, for all $j=1,\dots,d$, and for all values of $\rvx_{\pa_j}$, then $\vtheta_\alpha$ is a solution to 
 \[
 \argmin_{\vtheta\,:\, \P_{\rvx; \vtheta} \in \mathcal{D}} \operatorname{D}_{\mathrm{KL}}( \P_{\rvx ;\vtheta_p}  \,\|\, \P_{\rvx ; \vtheta}),
 \]
 where $\mathcal{D}$ denotes the set of SCMs that are Markov with respect to $\mathcal{G}_\alpha.$
\end{corollary}

\begin{proof}
 Because $\mathcal{G}_\alpha$ is a subgraph of $\mathcal{G}_p$, any SCM that is Markov relative to $\mathcal{G}_\alpha$ is Markov relative to $\mathcal{G}_p$. In particular,  Lemma~\ref{lem:factorize_KL} applies and gives that
\[
\operatorname{D}_{\mathrm{KL}}( \P_{\rvx ;\vtheta_p}  \,\|\, \P_{\rvx ; \vtheta_\alpha}) \leq \operatorname{D}_{\mathrm{KL}}( \P_{\rvx ;\vtheta_p}  \,\|\, \P_{\rvx ; \vtheta}).
\]
\end{proof}

In the case of linear Gaussian SCMs, Theorem~\ref{thm:optimal_adversarial_linear_gaussian_SCM} states that when $\mathcal{G}_\alpha$ is a subgraph of $\mathcal{G}_p$, we can find the optimal adversarial SCM in a closed form, in terms of the parameters of the true SCM.

\begin{theorem}[Optimal Adversarial SCM for linear Gaussian SCMs]
\label{thm:optimal_adversarial_linear_gaussian_SCM}
    Let $\P_{\rvx; \vtheta_p}$ be a mean-zero linear Gaussian SCM that is Markov with respect to $\mathcal{G}_p$, so $\rvx; \vtheta_p \sim \mathcal{N}(\bm{0}, \bm{\Sigma})$, and let $\mathcal{G}_\alpha$ be a subgraph of $\mathcal{G}_p.$ Denote the parents of $\rvx_j$ in $\mathcal{G}_\alpha$ by $\overline{\pa}_j$ and the parents of $\rvx_j$ in $\mathcal{G}_p$ by $\pa_j$. 
    
    The linear Gaussian SCM closest to $\P_{\rvx; \vtheta_p}$ in KL-divergence that is Markov with respect to $\mathcal{G}_\alpha$ is given by 
\[
 \mathbf{X} = \mB_\alpha^T \mathbf{X} + \vn,
 \]
 where $\vn \sim \mathcal{N}(\bm{0}, \mD)$ with
 \[
  \mD = \operatorname{Diag}\left(\left\{\bm{\Sigma}_{j,j} - \left \langle\bm{\Sigma}_{\overline{\pa}_j,j}, \left(\bm{\Sigma}_{\overline{\pa}_j, \overline{\pa}_j}\right)^{-1} \bm{\Sigma}_{\overline{\pa}_j,j}\right\rangle\right\}_{j=1}^d\right),
 \]
 and the $j$th column of $\mB$ is the vector 
 \[
 \vb_j = \mR_{\overline{\pa}_j}^T \left(\bm{\Sigma}_{\overline{\pa}_j, \overline{\pa}_j}\right)^{-1} \bm{\Sigma}_{\overline{\pa}_j,j},
 \]
where $\mR_{\overline{\pa}_j} \in \mathbb{R}^{|\overline{\pa}_j| \times d}$ is the restriction operator that satisfies $\mR_{\overline{\pa}_j} \vx = \vx_{\overline{\pa}_j}.$
\end{theorem}

\begin{proof}[Proof of Theorem~\ref{thm:optimal_adversarial_linear_gaussian_SCM}]
Let $\P_{\rvx; \vtheta}$ be Markovian with respect to $\mathcal{G}_\alpha$. Fix a $j$ in $1,\ldots,d$ and observe that the expected KL-divergence between $\P_{\rvx_j\mid \rvx_{\pa_j}; \vtheta_p}$ and $\P_{\rvx_j \mid \rvx_{\pa_j}; \vtheta}$ can be written in terms of an expected entropy and an expected cross-entropy term:
\begin{align*}
    \E_{\rvx_{\pa_j};\vtheta_p} \left[ \operatorname{D}_{\mathrm{KL}}\left( \P_{\rvx_j\mid \rvx_{\pa_j}; \vtheta_p} \,\|\, \P_{\rvx_j \mid \rvx_{\pa_j}; \vtheta} \right) \right] & \\
    & \kern -14em = \E_{\rvx_{\pa_j}; \vtheta_p} \left[ \E_{\rvx_j \mid \rvx_{\pa_j}; \vtheta_p} \left[ \ln \left( \frac{\p(\vx_j \mid \vx_{\pa_j}; \vtheta_p)}{\p(\vx_j \mid \vx_{\pa_j} ; \vtheta)}\right) \right] \right] \\
    & \kern -14em = \E_{\rvx_{\pa_j}; \vtheta_p} \left[ \E_{\rvx_j \mid \rvx_{\pa_j}; \vtheta_p} \left[ \ln ( \p(\vx_j \mid \vx_{\pa_j}; \vtheta_p) ) \right] \right] + \\
    & \kern -12em \E_{\rvx_{\pa_j}; \vtheta_p} \left[ \E_{\rvx_j \mid \rvx_{\pa_j}; \vtheta_p} \left[ - \ln( \p(\vx_j \mid \vx_{\pa_j} ; \vtheta) ) \right] \right]. \\
\intertext{Because the entropy term does not depend on $\vtheta$, we denote it with a constant $C_j$, and use the tower property of conditional expectation to continue further: }
    & \kern -14em = C_j + \E_{\rvx_{\pa_j \cup \{j\}}; \vtheta_p} \left[ -\ln( \p(\vx_j \mid \vx_{\pa_j} ; \vtheta) ) \right] \\
    & \kern -14em = C_j + \E_{\rvx_{\overline{\pa}_j \cup \{j\}; \vtheta_p}} \left[ -\ln( \p(\vx_j \mid \vx_{\overline{\pa}_j} ; \vtheta) ) \right] \\
    & \kern -14em = C_j + \E_{\rvx_{\overline{\pa}_j}; \vtheta_p} \left[ \E_{\rvx_j \mid \rvx_{\overline{\pa}_j}; \vtheta_p} \left[ - \ln( \p(\vx_j \mid \vx_{\overline{\pa}_j} ; \vtheta) ) \right] \right].
\intertext{ Here, the second equation follows from observing that the integrand depends only on the variables in $\overline{\pa}_j$, and the others are integrated out. Now we reverse the procedure to obtain the expectation of the KL-divergence between $\P_{\rvx_j\mid \rvx_{\overline{\pa}_j}; \vtheta_p}$ and $\P_{\rvx_j \mid \rvx_{\overline{\pa}_j}; \vtheta}$:}
    & \kern -14em = C_j - D_j + \E_{\rvx_{\overline{\pa}_j};\vtheta_p} \left[ \operatorname{D}_{\mathrm{KL}}\left( \P_{\rvx_j\mid \rvx_{\overline{\pa}_j}; \vtheta_p} \,\|\, \P_{\rvx_j \mid \rvx_{\overline{\pa}_j}; \vtheta} \right) \right],
\end{align*}
where $D = \E_{\rvx_{\overline{\pa}_j}; \vtheta_p} \left[ \E_{\rvx_j \mid \rvx_{\overline{\pa}_j}; \vtheta_p} \left[ \ln ( \p(\vx_j \mid \vx_{\overline{\pa}_j}; \vtheta_p) ) \right] \right]$ is a cross-entropy and does not depend on $\vtheta.$ 

As a consequence, we have established that
\begin{multline}
\label{eqn:sum_of_kl}
\sum_{j=1}^d \E_{\rvx_{\pa_j};\vtheta_p} \left[ \operatorname{D}_{\mathrm{KL}}\left( \P_{\rvx_j\mid \rvx_{\pa_j}; \vtheta_p} \,\|\, \P_{\rvx_j \mid \rvx_{\pa_j}; \vtheta} \right) \right] = \\
C - D + \sum_{j=1}^d \E_{\rvx_{\overline{\pa}_j};\vtheta_p} \left[ \operatorname{D}_{\mathrm{KL}}\left( \P_{\rvx_j\mid \rvx_{\overline{\pa}_j}; \vtheta_p} \,\|\, \P_{\rvx_j \mid \rvx_{\overline{\pa}_j}; \vtheta} \right) \right],
\end{multline}
where $C$ and $D$ do not depend on $\vtheta$. Now we explicitly construct an adversarial $\vtheta$ which minimizes the right-hand side.

 Since $\P_{\rvx; \vtheta_p}$ is a linear Gaussian SCM with mean zero and covariance $\bm{\Sigma}$, the posterior conditional $\rvx_j \mid \rvx_{\overline{\pa}_j}; \vtheta_p$ is Gaussian~\cite{murphy2012machine},
 \[ 
 \rvx_j \mid \rvx_{\overline{\pa}_j}; \vtheta_p \sim \mathcal{N}(\mu_{j\mid \overline{\pa}_k}, \sigma^2_{j\mid \overline{\pa}_j}),
 \]
 with mean and variance
 \begin{align*}
     \mu_{j\mid \overline{\pa}_k} & = \left\langle\left(\bm{\Sigma}_{\overline{\pa}_j, \overline{\pa}_j}\right)^{-1} \bm{\Sigma}_{\overline{\pa}_j,j}, \rvx_{\overline{\pa}_j} \right\rangle, \text{ and } \\  
     \sigma^2_{j\mid \overline{\pa}_j} & = \bm{\Sigma}_{j,j} - \left \langle\bm{\Sigma}_{\overline{\pa}_j,j}, \left(\bm{\Sigma}_{\overline{\pa}_j, \overline{\pa}_j}\right)^{-1} \bm{\Sigma}_{\overline{\pa}_j,j}\right\rangle.
 \end{align*}
 Accordingly, when $\P_{\rvx; \vtheta_\alpha}$ is selected as the linear gaussian SCM satisfying the conditions in the statement of the theorem, (1) for each $j$, $\P_{\rvx_j \mid \rvx_{\overline{\pa}_j}; \vtheta_\alpha}$ is exactly $\P_{\rvx_j \mid \rvx_{\overline{\pa}_j}; \vtheta_p}$, and (2) $\P_{\rvx; \vtheta_\alpha}$ is Markov with respect to $\mathcal{G}_\alpha$.

This choice of $\vtheta_\alpha$ minimizes the right-hand side of~\eqref{eqn:sum_of_kl} with respect to $\vtheta$, so by Corollary~\ref{cor:adversarial_optimality_condition}, $\vtheta_\alpha$ solves 
 \[
 \argmin_{\vtheta\,:\, \P_{\rvx; \vtheta} \in \mathcal{D}} \operatorname{D}_{\mathrm{KL}}( \P_{\rvx ;\vtheta_p}  \,\|\, \P_{\rvx ; \vtheta}).
 \]
\end{proof}

\section{Rejection Sampling for Adversarial Missingness}
\label{sxn:proofs_generalized_rejection_sampling}

The intuition behind the rejection sampling approach is to employ a missingness mechanism that biases towards choosing the observation pattern so that the observed features are more probable under $\P_{\rvx_o}(\cdot; \vtheta_\alpha)$ than under $P_{\rvx_o}(\cdot; \vtheta_p).$ Specifically, let $\vr \in \{0,1\}^d$ be a fixed nonzero observation mask, then define the ratio of the two probabilities for the corresponding observed features as 
\[
\Lambda_\vr(\vx_o) := \frac{\p_{\rvx_o}(\vx_o; \vtheta_\alpha)}{\p_{\rvx_o}(\vx_o; \vtheta_p)}
\]
and take $\Lambda_\vr^\star = \max_{\vx_o} \Lambda_\vr(\vx_o)$ to be the largest value of this ratio; we assume that the adversarial SCM is chosen so that $\Lambda_\vr^\star$ is finite. Now define the probability that the adversary chooses $\rvr = \vr$, conditional on observing the sample $\vx$ from $\P_{\rvx;\vtheta_p}$, in terms of these quantities,
\begin{equation}\label{eqn:generalized_rejection_sampler}
\p_{\rvr|\rvx}(\vr|\vx) = \begin{cases}
\pi_{\vr} \frac{\Lambda_\vr(\vx_o)}{\Lambda_\vr^\star} & \text{ if } \vr \neq 0 \\
1 - \sum_{\vr^\prime \neq 0} \pi_{\vr^\prime} \frac{\Lambda_{\vr^\prime}(\vx_{o^\prime})}{\Lambda_{\vr^\prime}^\star} & \text{ if } \vr = 0
\end{cases}.
\end{equation}
Here, the indices $o^\prime$ are defined with respect to $\vr^\prime$ and the weights $\pi_{\vr}$ satisfy $0 \leq \pi_{\vr} \leq 1$ for all $\vr \neq 0$. 

If the $\pi_{\vr}$ are selected so that the quantity defining the probability of $\rvr = 0$ conditional on $\rvx=\vx$ is nonnegative for all values of $\vx$, then the proposed missingness mechanism is a valid distribution. The choice $\pi_{\vr} = \frac{1}{2^d - 1}$, which assigns uniform weight to all nonzero observation patterns, is always valid, independent of the choice of the adversarial and true distributions; this is true because $\Lambda_{\vr}/\Lambda_{\vr}^\star \leq 1$ always, for all non-zero observation patterns $\vr$.

It is worth noting that for this mechanism, the MAR assumption fails only at $\vr = 0$, as otherwise $\rvr$ depends only on the observed features $\vx_o$, so $\rvx_m \perp \rvr = \vr \,|\, \rvx_o$. 

Our first observation is that this generalized rejection sampling approach ensures that the partially observed features from $\P_{\rvx;\vtheta_p}$ look as though they were sampled from the adversarial distribution.

\begin{lemma}[Generalized Rejection Sampling]\label{lem:generalized_rejection_sampling}
	The distribution over the observed features induced from $\P_{\rvx; \vtheta_p}$ by the missingness mechanism defined in~\eqref{eqn:generalized_rejection_sampler} satisfies
 \[\P_{\rvx_o \mid \rvr}(\cdot\,| \vr; \vtheta_p)=\P_{\rvx_o}(\cdot \,;\vtheta_\alpha)\]
 for all $\vr\neq 0$.
\end{lemma}

\begin{proof}
Fix an observation pattern $\vr \neq 0$ and the corresponding subset of observed features $\vx_o$. Observe that
\begin{align*}
\p_{\rvx_o \mid \rvr} (\vx_o \mid \vr) & = \frac{\p_{\rvr | \rvx_o}(\rv | \vx_o) \p_{\rvx_o}(\vx_o ; \vtheta_p) }{\p_{\rvr}(\vr)} \\
 & = \frac{\pi_\vr \Lambda_\vr(\vx_o)}{\p_\rvr(\vr) \Lambda_\vr^\star} \p_{\rvx_o}(\vx_o; \vtheta_p) \\
 & = \frac{\pi_\vr}{\p_\rvr(\vr) \Lambda_\vr^\star}\p_{\rvx_o}(\vx_o; \vtheta_\alpha) \\
 & \propto \p_{\rvx_o}(\vx_o; \vtheta_\alpha).
\end{align*}
The first equality uses Bayes' Theorem, the second uses the definition of the missingness mechanism for generalized rejection sampling (\eqref{eqn:generalized_rejection_sampler}), and the third uses the definition of $\Lambda_\vr(\vx_o).$ We have shown that, for each observation pattern $\vr \neq 0$, the density functions $\p_{\rvx_o \mid \rvr} (\cdot \mid \vr)$ and $\p_{\rvx_o}( \cdot \,; \vtheta_\alpha)$ are proportional, therefore the distributions are in fact equal, as claimed.
\end{proof}

When the missingness mechanism is such that the partially observed features for any observation pattern are distributed as though they were drawn from the adversarial distribution, the adversary is able to manipulate the modeler into thinking the observed data comes from the adversarial distribution.

\begin{theorem}[Global Optimality of the Adversarial SCM]\label{thm:rejection_sampling_global_maximizer}
When the adversarial missingness mechanism $\P_{\rvr|\rvx}$ satisfies the condition
\[\P_{\rvx_o \mid \rvr}(\cdot\,| \vr; \vtheta_p)=\P_{\rvx_o}(\cdot \,;\vtheta_\alpha)\]
for all $\vr\neq 0$, it is the case that $\vtheta_\alpha$ is a global maximizer of the modeler's objective (\eqref{eqn:modeler-objective}).
\end{theorem}

\begin{proof}
Recall that the modeler's objective (\eqref{eqn:modeler-objective}) is 
\[
J(\vtheta) = \E_{\rvr|\rvr\neq 0}\left[ \E_{\rvx_o|\rvr;\vtheta_p}\!\left[ \log \p_{\rvx_o;\vtheta}(\vx_o) \mid \rvr = \vr\right] \right].
\]
Fix a parameter $\vtheta \in \mathcal{D}$ and compute
\begin{align*}
J(\vtheta_\alpha)  - J(\vtheta) & =  \sum_{\vr \neq 0} \p_{\rvr\mid\rvr \neq 0}(\vr) \E_{\rvx_o\mid \rvr}\left[ \log \frac{\p_{\rvx_o}(\vx_o;\vtheta_\alpha)}{\p_{\rvx_o}(\vx_o; \vtheta)}\,\bigg|\, \rvr = \vr \right] \\
& = \sum_{\vr \neq 0} \p_{\rvr\mid\rvr \neq 0}(\vr) \E_{\rvx_o; \vtheta_\alpha}\left[ \log \frac{\p_{\rvx_o}(\vx_o;\vtheta_\alpha)}{\p_{\rvx_o}(\vx_o; \vtheta)}\,\bigg|\, \rvr = \vr \right] \\
& = \sum_{\vr \neq 0} \p_{\rvr\mid \rvr \neq 0}(\vr) \textrm{D}_{\textrm{KL}}(\P_{\rvx_o}(\cdot\,; \vtheta_\alpha)\,\|\, \P_{\rvx_o}(\cdot\,; \vtheta)) \\
& \geq 0.
\end{align*}
The second equality is justified by Lemma~\ref{lem:generalized_rejection_sampling}. This inequality shows that $\vtheta_\alpha$ is a global maximizer of the modeler's objective.
\end{proof}

It follows immediately that the adversarial model is a global maximizer of the modeler's objection~\eqref{eqn:modeler-objective}. 

\begin{corollary}\label{cor:attack_success_generalized_rejection_sampling}
When the adversarial missingness distribution is constructed using the generalized rejection sampling of~\eqref{eqn:generalized_rejection_sampler}, $\vtheta_\alpha$ is a global maximizer of the modeler's objective (\eqref{eqn:modeler-objective}).
\end{corollary}

\begin{proof}
When the missingness distribution is constructed using the generalized rejection sampling of~\eqref{eqn:generalized_rejection_sampler}, Lemma~\ref{lem:generalized_rejection_sampling} states that 
the condition
\[\P_{\rvx_o \mid \rvr}(\cdot\,| \vr; \vtheta_p)=\P_{\rvx_o}(\cdot \,;\vtheta_\alpha)\]
is satisfied for all $\vr\neq 0$. Consequently, Theorem~\ref{thm:rejection_sampling_global_maximizer} applies and yields that $\vtheta_\alpha$ is a global maximizer of the modeler's objective, as claimed.
\end{proof}

This result shows that generalized rejection sampling achieves several of the goals of the adversary stipulated in Section~\ref{sxn:formulation}. Specifically, if one assumes that the adversary has chosen an adversarial distribution that is $\beta$-indistinguishable and satisfies adversarial Markovianity, then Corollary~\ref{cor:attack_success_generalized_rejection_sampling} states that the attack succeeds in the sense that the adversarial model is a maximizer of the modeler's objective. As the modeler's objective is nonconvex and may have multiple global optima, this seems likely to be the strongest type of result one can obtain on attack success without restricting the class of SCMs under consideration.

A bound on the missingness rate for the generalized rejection sampling missingness mechanism can be given. Recall that for the missingness mechanism in~\eqref{eqn:generalized_rejection_sampler} to be well-defined, the weights $\pi_\vr$ must be selected to ensure that the probability of sampling $\rvr = 0$ given $\rvx = \vx$ is nonnegative for any $\vx$. The choice of weights also affects the bound on the missingness rate.

\begin{lemma}\label{lem:generalized_rejection_sampling_missingness_bound}
Let $\ell_{\vr} = |\{j \mid \vr_j = 0\}|$ denote the amount of features missing in pattern $\vr$. When generalized rejection sampling is used, the expected missingness rate is given by
\[
\E_{\rvr}\left[\frac{|\{j\,|\,\rvr_j=0\}|}{d}\right] = 1 - \sum_{\vr \neq 0} \frac{\pi_\vr}{\Lambda_\vr^\star} \left(1 - \frac{\ell_{\vr}}{d} \right)
\].
\end{lemma}

Lemma~\ref{lem:generalized_rejection_sampling_missingness_bound} shows that the choice of $\pi_{\vr}$ has an important effect on the missingness rate. Specifically, it establishes the importance of employing large weights $\pi_{\vr}$ in order to reduce the expected amount of missing data. It is difficult to select admissible and large $\pi_{\vr}$, as the weights must be chosen so that the probability of sampling $\rvr = 0$ conditional on \emph{any} sample $\rvx = \vx$ is a valid probability. 

\begin{proof}[Proof of Lemma~\ref{lem:generalized_rejection_sampling_missingness_bound}]
For all $\vr \neq 0$, it is the case that
\begin{align*}
\p_{\rvr}(\vr) & = \E_{\rvx;\vtheta_p} \left[ \p_{\rvr | \rvx}(\vr \mid \vx) \right] = \E_{\rvx;\vtheta_p} \left[ \pi_{\vr} \frac{\Lambda_\vr(\vx_o)}{\Lambda_\vr^\star} \right] \\
& = \frac{\pi_{\vr}}{\Lambda_{\vr}^\star} \E_{\rvx_o;\vtheta_p} \left[ \frac{\p_{\rvx_o}(\vx_o; \vtheta_\alpha)}{\p_{\rvx_o}(\vx_o; \vtheta_p)} \right] = \frac{\pi_{\vr}}{\Lambda_{\vr}^\star}.
\end{align*}
The first equality is the law of total probability, the second and third are justified by the definition of the missingness mechanism, and the forth follows from the fact that pdfs integrate to one.

Now the expected missingness rate can be computed as 
\begin{align*}
 \frac{1}{d}\E_{\rvr}[\ell_{\vr}] & = \p_{\rvr}(0) \frac{\ell_0}{d} + \frac{1}{d} \sum_{\vr \neq 0} \p_{\rvr}(\vr) \ell_{\vr} \\
  & = \left(1 - \sum_{\vr\neq 0} \p_{\rvr}(\vr) \right) + \frac{1}{d} \sum_{\vr\neq 0} \frac{\pi_{\vr}}{\Lambda_{\vr}^\star} \ell_{\vr} \\
  & = 1 - \sum_{\vr \neq 0} \frac{\pi_\vr}{\Lambda_\vr^\star} \left(1 - \frac{\ell_{\vr}}{d} \right).
\end{align*}
The first equality is an expansion of the expectation, the second uses the fact that $\ell_0 = d$ and the expression found for $\p_{\rvr}$, while the final equality is justified by algebra.
\end{proof}

As noted earlier, the choice $\pi_{\vr} = (2^d - 1)^{-1}$ is \emph{always} admissible. However, in cases where the adversarial SCM is close to the true SCM, one can find missingness mechanisms with significantly tighter bounds on their missingness rate by utilizing \emph{localized rejection sampling} as described in Section~\ref{sxn:generalized_rejection_sampling}.

Recall the definition of localized rejection sampling. Let $\mathcal{V} \subseteq \{1, \ldots, d\}$ denote a subset of the variables, and $\overline{\mathcal{V}}$ denote the complement. Localized generalized rejection sampling on the variables $\mathcal{V}$ is a missingness mechanism that masks only variables in $\mathcal{V}$, using probabilities depending only on the value of $\rvx_{\mathcal{V}}$, given by

\begin{equation}
\P_{\rvr|\rvx}(\vr|\vx) = \begin{cases}
\frac{1}{2^{|\mathcal{V}|}-1} \frac{\Lambda(\vx_{\mathcal{V}})}{\Lambda} & \text{ if } \vr_{\overline{\mathcal{V}}}=1 \text{ and } \vr_{\mathcal{V}} \neq 0 \\
1 - \frac{\Lambda(\vx_{\mathcal{V}})}{\Lambda} & \text{ if } \vr_{\overline{\mathcal{V}}}=1 \text{ and } \vr_{\mathcal{V}} = 0 \\
0 & \text{ otherwise}
\end{cases}.
\end{equation}

Here, $\Lambda(\vx_{\mathcal{V}}) = \frac{\p_{\rvx_{\mathcal{V}};\vtheta_\alpha}(\vx_{\mathcal{V}})} {\p_{\rvx_{\mathcal{V}};\vtheta_p}(\vx_{\mathcal{V}})} $ is the ratio of the adversarial distribution to the true distribution, and $\Lambda = \max_{\vx_{\mathcal{V}}} \Lambda(\vx_{\mathcal{V}})$ is the maximum value of that ratio. Note that the observation patterns that select all variables in $\overline{\mathcal{V}}$ and at least one variable in $\mathcal{V}$ are equiprobable. Because this approach only drops variables in $\mathcal{V}$, the missingness rate is at most $\tfrac{|\mathcal{V}|}{d}$, which is significantly better than the worst-case $O(1)$ missingness rate of the general rejection sampling approach given earlier. 

When the conditional distributions of the variables in $\overline{\mathcal{V}}$ given the variables in $\mathcal{V}$ is identical in the adversarial and true SCMs, localized generalized rejection sampling ensures that the partially observed features from $\P_{\rvx;\vtheta_p}$ look as though they were sampled from the adversarial distribution.

\localizedrejectionsampling*

\begin{proof}
Fix an arbitrary $\vr$ satisfying $\p_\rvr(\vr) \neq 0$, and note that this implies that $\vr_{\overline{\mathcal{V}}}=1$. Fix a corresponding subset of observed features $\vx_o$. Observe that 
\begin{equation*}
\p_{\rvx_o\mid \rvr}(\vx_o \mid \vr; \vtheta_p) = \sum_{\vx_m} \p_{\rvx \mid \rvr}(\vx \mid \vr; \vtheta_p),
\end{equation*}
where the summation with respect to $\vx_m$ denotes summation over all the possible values of the unobserved features. It follows that 
\begin{align*}
\p_{\rvx_o\mid \rvr}(\vx_o \mid \vr; \vtheta_p) & = \sum_{\vx_m} \p_{\rvx \mid \rvr}(\vx \mid \vr; \vtheta_p) \\
& = \sum_{\vx_m} \p_{\rvx_{\overline{\mathcal{V}} \mid \mathcal{V}, \rvr}} (\vx_{\overline{\mathcal{V}}} \mid \vx_{\mathcal{V}}, \vr; \vtheta_p) \p_{\rvx_\mathcal{V} \mid \rvr}(\vx_{\mathcal{V}}\mid \vr; \vtheta_p) \\
& = \sum_{\vx_m} \p_{\rvx_{\overline{\mathcal{V}} \mid \mathcal{V}}} (\vx_{\overline{\mathcal{V}}} \mid \vx_{\mathcal{V}}; \vtheta_p) \p_{\rvx_\mathcal{V} \mid \rvr}(\vx_{\mathcal{V}}\mid \vr; \vtheta_p),
\end{align*}
where the last two equalities are justified by the chain rule and the fact that $\rvr \independent \rvx_{\overline{\mathcal{V}}} \mid \rvx_{\mathcal{V}}$. Now we apply the assumption that
\[
\P_{\rvx_{\overline{\mathcal{V}}} \mid \rvx_{\mathcal{V}}}( \cdot \mid  \cdot\,; \vtheta_\alpha) = \P_{\rvx_{\overline{\mathcal{V}}} \mid \rvx_{\mathcal{V}}}(\cdot \mid \cdot\, ; \vtheta_p),
\]
to see that
\begin{equation}\label{eqn:01}
\p_{\rvx_o\mid \rvr}(\vx_o \mid \vr; \vtheta_p) = \sum_{\vx_m} \p_{\rvx_{\overline{\mathcal{V}} \mid \mathcal{V}}} (\vx_{\overline{\mathcal{V}}} \mid \vx_{\mathcal{V}}; \vtheta_\alpha) \p_{\rvx_\mathcal{V} \mid \rvr}(\vx_{\mathcal{V}}\mid \vr; \vtheta_p).
\end{equation}

In fact, $\p_{\rvx_{\mathcal{V}} \mid \rvr }(\vx_{\mathcal{V}} \mid \vr; \vtheta_p) = \p_{\rvx_\mathcal{V}}(\vx_{\mathcal{V}}; \vtheta_\alpha)$. We use similar manipulations as those in the proof of Lemma~\ref{lem:generalized_rejection_sampling} to establish this fact. First, consider the case that $\vr_{\mathcal{V}} \neq 0$: 
\begin{align*}
\p_{\rvx_{\mathcal{V}} \mid \rvr }(\vx_{\mathcal{V}} \mid \vr; \vtheta_p) & = 
\frac{\p_{\rvr \mid \rvx_{\mathcal{V}}}(\rv \mid \vx_{\mathcal{V}}) \p_{\rvx_{\mathcal{V}}}(\vx_\mathcal{V}; \vtheta_p)}{\p_{\rvr}(\vr)} \\
& = \frac{1}{(2^{|\mathcal{V}|}-1) \Lambda \p_{\rvr}(\vr)} \Lambda(\vx_{\mathcal{V}}) \p_{\rvx_{\mathcal{V}}}(\vx_{\mathcal{V}};\vtheta_p) \\
& = \frac{1}{(2^{|\mathcal{V}|}-1) \Lambda \p_{\rvr}(\vr)} \p_{\rvx_\mathcal{V}}(\vx_\mathcal{V};\vtheta_\alpha) \\
& \propto \p_{\rvx_\mathcal{V}}(\vx_\mathcal{V};\vtheta_\alpha).
\end{align*}
The first equation is Bayes' Theorem, the second follows from the definition of the localized generalized rejection sampling missingness mechanism (\eqref{eqn:localized_generalized_rejection_sampling}), and the third uses the definition of $\Lambda(\vx_{\mathcal{V}})$. Since the pdfs $\p_{\rvx_{\mathcal{V}} \mid \rvr }( \cdot \mid \vr; \vtheta_p)$ and $\p_{\rvx_\mathcal{V}}(\cdot\,;\vtheta_\alpha)$ are proportional, they are in fact equal in the case that $\vr_{\mathcal{V}} \neq 0$. 

Similarly, when $\vr_{\mathcal{V}} = 0$, we have that
\begin{align*}
\p_{\rvx_{\mathcal{V}} \mid \rvr }(\vx_{\mathcal{V}} \mid \vr; \vtheta_p) & = \frac{1}{\p_{\rvr}(\vr)} \left(1 - \frac{\Lambda(\vx_{\mathcal{V}})}{\Lambda}\right) \p_{\rvx_{\mathcal{V}}}(\vx_{\mathcal{V}}; \vtheta_p) \\
& = \frac{1}{\p_{\rvr}(\vr)} \left( \p_{\rvx_{\mathcal{V}}}(\vx_{\mathcal{V}}; \vtheta_p) - \frac{1}{\Lambda} \p_{\rvx_{\mathcal{V}}}(\vx_{\mathcal{V}}; \vtheta_\alpha) \right),
\end{align*}
where the second equality follows from the definition of $\Lambda(\vx_{\mathcal{V}})$. Equivalently,
\begin{equation}\label{eqn:02}
\p_{\rvr}(\vr) \p_{\rvx_{\mathcal{V}} \mid \rvr }(\vx_{\mathcal{V}} \mid \vr; \vtheta_p) = \p_{\rvx_{\mathcal{V}}}(\vx_{\mathcal{V}}; \vtheta_p) - \frac{1}{\Lambda} \p_{\rvx_{\mathcal{V}}}(\vx_{\mathcal{V}}; \vtheta_\alpha).
\end{equation}
Observe that when $\vr_{\mathcal{V}}= 0$, by the definition of the localized generalized rejection sampling missingness mechanism,
\begin{equation*}
\p_{\rvr}(\vr) = \E_{\rvx;\vtheta_p} \p_{\rvr\mid\rvx}(\vr\mid\vx) 
 = \E_{\rvx;\vtheta_p} \left( 1 - \frac{\Lambda(\vx_{\mathcal{V}})}{\Lambda} \right)  = 1 - \frac{1}{\Lambda},
\end{equation*}
where the last equation holds because $\E_{\rvx;\vtheta_p} \Lambda(\vx_{\mathcal{V}}) = 1.$ It follows from this observation and~\eqref{eqn:02} that $\p_{\rvx_{\mathcal{V}} \mid \rvr }(\vx_{\mathcal{V}} \mid \vr; \vtheta_p) = \p_{\rvx_{\mathcal{V}}}(\vx_{\mathcal{V}}; \vtheta_\alpha)$ when $\vr_{\mathcal{V}} = 0.$
Thus, as claimed, we have established that for any arbitrary $\vr$ satisfying $\p_\rvr(\vr) \neq 0$, it is the case that $\p_{\rvx_{\mathcal{V}} \mid \rvr }(\vx_{\mathcal{V}} \mid \vr; \vtheta_p) = \p_{\rvx_\mathcal{V}}(\vx_{\mathcal{V}}; \vtheta_\alpha)$.

Returning to~\eqref{eqn:01} and using this fact, we have that as claimed,
\begin{align*}
\p_{\rvx_o\mid \rvr}(\vx_o \mid \vr; \vtheta_p) & = \sum_{\vx_m} \p_{\rvx_{\overline{\mathcal{V}} \mid \mathcal{V}}} (\vx_{\overline{\mathcal{V}}} \mid \vx_{\mathcal{V}}; \vtheta_\alpha)p_{\rvx_\mathcal{V}}(\vx_{\mathcal{V}}; \vtheta_\alpha) \\
& = \sum_{\vx_m} \p_{\rvx}(\vx; \vtheta_\alpha) \\
& = \p_{\rvx_o}(\vx_o; \vtheta_\alpha)
\end{align*}
for any $\vr$ satisfying $\p_\rvr(\vr) \neq 0.$
\end{proof}
This result implies that when the matching condition~\eqref{eqn:matched_conditional_distribution} holds, the adversary can attain their goal of causing $\vtheta_\alpha$ to be a global maximizer of the modeler's objective.

\attacksuccesslocalized*

\begin{proof}
When the assumed conditions hold, Lemma~\ref{lem:localized_generalized_rejection_sampling} applies, so~\eqref{eqn:matched_conditional_distribution} holds. Then Theorem~\ref{thm:rejection_sampling_global_maximizer} applies and says that $\vtheta_\alpha$ is a global maximizer of the objective of the modeler.
\end{proof}

As an example application of Corollary~\ref{cor:attack_success_localized_generalized_rejection_sampling}, consider the case in which $\mathcal{G}_\alpha$ is obtained from $\mathcal{G}$ by deleting some number of incoming edges to a node $s$ and the adversarial SCM is constructed by preserving the functional relationships between $j$ and its parents whenever $j \neq s$ and imposing some new functional relationship between $s$ and its reduced set of parents. Then it is the case that~\eqref{eqn:matched_conditional_distribution} holds, so that the adversarial distribution is a global maximizer of the modeler's objective when localized generalized rejection sampling is used with $\mathcal{V} = \{s\} \cup \pa_s$.

\begin{corollary}[Local Edge Deletion]
\label{cor:local_edge_deletion}
Distinguish between the parents of variable $j$ in $\mathcal{G}$ and in $\mathcal{G}_\alpha$ by denoting the former with $\pa_j$ and the latter with $\overline{\pa}_j$. Select a $\mathcal{G}_\alpha$ such that $\overline{\pa}_j$ = $\overline \pa_j$ for all $j \neq s$ and $\overline{pa}_s \subseteq \pa_s$. If the adversarial distribution satisfies
\[
\P_{\rvx_j\mid\rvx_{\pa_j}}(\cdot\mid\cdot\,;\vtheta_p) = \P_{\rvx_j\mid\rvx_{\pa_j}}(\cdot\mid\cdot\,;\vtheta_\alpha) \text{ when $j \neq s$},
\]
then~\eqref{eqn:matched_conditional_distribution} holds and it follows that when the adversary uses localized generalized rejection sampling (\eqref{eqn:localized_generalized_rejection_sampling}) with $\mathcal{V} = \{s\} \cup \pa_s$, then $\vtheta_\alpha$ is a global maximizer of the objective of the modeler.
\end{corollary}

\begin{proof}[Proof of Corollary~\ref{cor:local_edge_deletion}]
Since the edges in $\mathcal{G}_\alpha$ are a subset of the edges of $\mathcal{G}_p$ and $\P_{\rvx;\vtheta_\alpha}$ is Markov relative to $\mathcal{G}_\alpha$, it is also Markov relative to $\mathcal{G}_p$.

Use the causal factorization of $\P_{\rvx;\vtheta_\alpha}$ relative to $\mathcal{G}_p$ to calculate the pdf of $\rvx_{\overline{\mathcal{V}}}$ conditional on $\rvx_\mathcal{V}$ under the adversarial distribution,
\begin{equation*}
	\begin{split}
		 &\p_{\rvx_{\overline{\mathcal{V}}} \mid \rvx_\mathcal{V}}(\vx_{\overline{\mathcal{V}}} \mid \vx_\mathcal{V} ; \vtheta_p) = 
		 \frac{\jp}
		 {\sum_{\vx_{\fu}}\jp}\\
		&=\frac{\cpd}{\cpd}\frac{\fp}{\sum_{\vx_{\fu}}\fp}\\
		&=\frac{\fp}{\sum_{\vx_{\fu}}\fp}\\
		&=\frac{\fa}{\sum_{\vx_{\fu}}\fa}\\
		&=\frac{\cpda}{\cpda}\frac{\fa}{\sum_{\vx_{\fu}}\fa}\\
		&=\frac{\ja}{\sum_{\vx_{\fu}}\ja}\\	
		&=\p_{\rvx_{\overline{\mathcal{V}}} \mid \rvx_\mathcal{V}}(\vx_{\overline{\mathcal{V}}} \mid \vx_\mathcal{V} ; \ta) 
	\end{split}
\end{equation*}

The first equality follows from the definition of conditional probability and the law of total probability. The second follows from the causal factorization and the fact that the variables indexed by $\nfu=\{s\}\cup \pa_s$ are not in the summation. The third equality is simple algebra, and the fourth follows from the assumption about the equality of the conditional marginals of the adversarial and true distributions. The last equality follows from the definition of conditional probability and the law of total probability. 

Because $\P_{\rvx_{\overline{\mathcal{V}}} \mid \rvx_{\mathcal{V}}}( \cdot \mid  \cdot\,; \vtheta_\alpha) = \P_{\rvx_{\overline{\mathcal{V}}} \mid \rvx_{\mathcal{V}}}(\cdot \mid \cdot\, ; \vtheta_p),$ holds, Corollary \ref{cor:attack_success_localized_generalized_rejection_sampling} applies.
\end{proof}

As stated earlier, the missingness rate of localized rejection sampling is at most $\tfrac{|\mathcal{V}|}{d}$, which is significantly better than the worst-case $O(1)$ missingness rate of the general rejection sampling approach given earlier. In fact, a tighter bound can be found on the missingness rate. The following result shows that the missingness rate is smaller when the adversarial and true distribution are close in the sense that the maximum of their ratio, $\Lambda$, is small.

\begin{lemma}\label{lem:localized_generalized_rejection_sampling_missingness_bound}
Let $\ell_{\vr} = |\{j \mid \vr_j = 0\}|$ denote the amount of features missing in pattern $\vr$. When localized generalized rejection sampling is used on the variables $\mathcal{V}$, the expected missingness rate satisfies
\begin{align*}
\E_{\rvr}\left[\frac{|\{j\,|\,\rvr_j=0\}|}{d}\right]  & = 
 \left(1 - \frac{1}{\Lambda}\right)\frac{|\mathcal{V}|}{d} + \\
 & \kern -3em \frac{1}{\Lambda d }\cdot \frac{1}{\left(2^{|\mathcal{V}|} - 1\right)} \sum_{\vr:\vr_{\overline{\mathcal{V}}}=1, \vr_{\mathcal{V}} \neq \bm{0}} \ell_\vr,
\end{align*}
and consequently,
\begin{equation*}
\E_{\rvr}\left[\frac{|\{j\,|\,\rvr_j=0\}|}{d}\right] \leq \left(1 - \frac{1}{2 \Lambda} \right) \frac{|\mathcal{V}|}{d}.
\end{equation*}
\end{lemma}

\begin{proof}
To compute the expected missingness rate, first compute the probability of each observation pattern satisfying $\vr_{\overline{\mathcal{V}}}=\bm{1}$ and $\vr_{\mathcal{V}}\neq \bm{0}$,
\begin{align*}
    \p_{\rvr}(\vr) & = \E_{\rvx; \vtheta_p}[ \p_{\rvr\mid\rvx}(\vr\mid\vx)] = \E_{\rvx; \vtheta_p}\left[ \frac{\Lambda(\vx_\mathcal{V})}{\Lambda} \frac{1}{2^{|\mathcal{V}|} - 1} \right]  \\
    & = = \frac{1}{(2^{|\mathcal{V}|} - 1) \Lambda} \E_{\rvx_{\mathcal{V};\vtheta_p}}[ \Lambda(\vx_\mathcal{V})] \\
    & = \frac{1}{(2^{|\mathcal{V}|} - 1) \Lambda} \E_{\rvx_{\mathcal{V};\vtheta_p}}\left[ \frac{\p_{\rvx_{\mathcal{V}}}(\vx_{\mathcal{V}}; \vtheta_\alpha)}{\p_{\rvx_{\mathcal{V}}}(\vx_{\mathcal{V}}; \vtheta_p)} \right] 
    =  \frac{1}{(2^{|\mathcal{V}|} - 1) \Lambda}.
\end{align*}
When $\vr_{\overline{\mathcal{V}}}=\bm{1}$ and $\vr_{\mathcal{V}}= \bm{0}$, then similarly
\begin{align*}
    \p_{\rvr}(\vr) & = \E_{\rvx; \vtheta_p}[ \p_{\rvr\mid\rvx}(\vr\mid\vx)] = \E_{\rvx; \vtheta_p}\left[ 1 - \frac{\Lambda(\vx_\mathcal{V})}{\Lambda} \right]  \\
    & = = 1 - \frac{1}{\Lambda} \E_{\rvx_{\mathcal{V};\vtheta_p}}[ \Lambda(\vx_\mathcal{V})]\\
    & = 1 - \frac{1}{\Lambda} \E_{\rvx_{\mathcal{V};\vtheta_p}}\left[ \frac{\p_{\rvx_{\mathcal{V}}}(\vx_{\mathcal{V}}; \vtheta_\alpha)}{\p_{\rvx_{\mathcal{V}}}(\vx_{\mathcal{V}}; \vtheta_p)} \right] 
    = 1 - \frac{1}{\Lambda}.
\end{align*}
Thus the expected missingness rate can be calculated as 
\begin{equation*}
    \E_{\rvr}\left[\frac{|\{j\,|\,\rvr_j=0\}|}{d}\right] = \p_{\rvr}(\bm{0}) \frac{|\mathcal{V}|}{d} +  
    \frac{1}{d} \sum_{\vr\,:\, \vr_{\overline{\mathcal{V}}} = \bm{1}, \vr_{\mathcal{V}} \neq \bm{0} } \p_{\rvr}(\vr) \E [ \ell_\vr],
\end{equation*}
where $\ell_{\rv}$ is defined as in Lemma~\ref{lem:generalized_rejection_sampling_missingness_bound}. Inserting the expressions for $\p(\vr)$ from above establishes the first claim of this lemma,
\begin{align}
\label{eqn:first_claim}
\E_{\rvr}\left[\frac{|\{j\,|\,\rvr_j=0\}|}{d}\right] & = 
 \left(1 - \frac{1}{\Lambda}\right)\frac{|\mathcal{V}|}{d} + \\
 & \kern -3em \frac{1}{\Lambda d }\cdot \frac{1}{\left(2^{|\mathcal{V}|} - 1\right)} \sum_{\vr:\vr_{\overline{\mathcal{V}}}=1, \vr_{\mathcal{V}} \neq \bm{0}} \ell_\vr.
\end{align}
The second claim follows from the observation that each coordinate in $\rv_{\mathcal{V}}$ can be viewed as the result of flipping a fair coin. This implies that the average of the $\ell_{\rv}$ on the right-hand side of this equation is the expected number of failures in $|\mathcal{V}|$ independent experiments, each with probability $\tfrac{1}{2}$ of success, conditioned on the fact that there is at least one success. That is,
\begin{align*}
\frac{1}{\left(2^{|\mathcal{V}|} - 1\right)} \sum_{\vr:\vr_{\overline{\mathcal{V}}}=1, \vr_{\mathcal{V}} \neq \bm{0}} \ell_\vr & = \E_{b \sim \operatorname{Bin}(|\mathcal{V}|, \frac{1}{2})}[|\mathcal{V}| -  b \mid b > 0] \\
& \kern -6em = |\mathcal{V}| - \frac{\E[b]}{\P(b \neq 0)} \leq |\mathcal{V}| - \E[b]
 = \frac{|\mathcal{V}|}{2}.
\end{align*}
Using this estimate in \eqref{eqn:first_claim} gives, as claimed,
\begin{align*}
    \E_{\rvr}\left[\frac{|\{j\,|\,\rvr_j=0\}|}{d}\right]  \leq 
 \left(1 - \frac{1}{\Lambda}\right)\frac{|\mathcal{V}|}{d} + \frac{1}{2} \frac{|\mathcal{V}|}{\Lambda d} = \left(1 - \frac{1}{2 \Lambda} \right) \frac{|\mathcal{V}|}{d}.
\end{align*}
\end{proof}

\section{The Weighted EM (WEM) Algorithm}\label{app:wem}

Denote the missingness weights by
\[
\omega_{i,r}(\vphi)=\frac{\P_{\rvr \mid \rvx} (\vr \mid \vx^{(i)} ; \bm{\phi})}{\sum_{\vr^\prime \neq 0} \P_{\rvr \mid \rvx } (\rv^\prime\mid \vx^{(i)}; \bm{\phi})},
\]
where $\data = \{\vx^{(i)}\}_{i=1}^N$ is the fully observed training data. Given the parameters $\vtheta^{t-1}$ from the previous step, the expectation step of WEM is the calculation of
\begin{equation}
	Q(\vtheta, \vtheta^{t-1},\bm{\phi}) = \sum_{i=1}^N \sum_{\vr\neq 0}\omega_{i,r}(\vphi)
	\E_{\rvx;\vtheta^{t-1}}\left[\log \P_{\rvx}(\rvx; \vtheta) \mid \rvx_{o}=\vx_{o}^{(i)}\right] \label{eqn:expectation_step_of_WEM}
\end{equation}
When $N$ is large, by the law of large numbers the expectation step of EM is close to the expectation step of WEM. The maximization step of WEM does not employ a DAG constraint, and is given by
\begin{equation}\label{eq:wem}
	\vtheta^t = \argmax_{\vtheta} Q(\vtheta, \vtheta^{t-1},\bm{\phi}).
\end{equation}

The WEM Algorithm, which repeats WEM expectation and maximization steps until convergence of the parameters $\vtheta^t$, is described in Algorithm~\ref{alg:wem}. It utilizes a likelihood-based stopping criterion. Specifically, an empirical approximation of the modeler's objective function in ~\eqref{eqn:modeler-objective} is used to measure the probability of the observed data; again, the expectation over the missingness mechanism is explicitly computed:
\begin{equation}
\label{eqn:empirical_likelihood}
J(\vtheta,\vphi) = \sum_{i=1}^N \sum_{r\neq 0} \omega_{i,r}(\vphi) \log \P_{\rvx_o}(\vx^{(i)}_o;\vtheta)
\end{equation}

\begin{algorithm}[H]
	\caption{WEM Algorithm. Solves the weighted EM problem corresponding to a given missingness mechanism, using a likelihood-based stopping criterion. Both the likelihood $J(\vtheta,\vphi)$ (\eqref{eqn:empirical_likelihood}) and the expectation $Q(\vtheta, \vtheta^{t-1},\bm{\phi})$ (\eqref{eqn:expectation_step_of_WEM}) use the fully observed samples, $\data$. This dependence is omitted from the notation for brevity.}
	\label{alg:wem}
	
	\begin{algorithmic}
		\STATE {\bfseries Input: $\vphi, \vtheta^{0}, \epsilon, \data$ }
		\STATE $t\leftarrow 1$
		\WHILE{$J(\vtheta^{(t)},\bm{\phi})-J(\vtheta^{(t-1)},\bm{\phi})\geq \epsilon |J(\vtheta^{(t-1)},\bm{\phi})|$} 
		\STATE $\vtheta^{t}= \argmax_{\vtheta} Q(\vtheta, \vtheta^{t-1},\bm{\phi}) $
		\STATE $t\leftarrow t+1$
		\ENDWHILE
		\STATE $\ttil \leftarrow \vtheta^{t-1}$
		\STATE {\bfseries Output: $\ttil$} 
	\end{algorithmic}
\end{algorithm}

We have provided the WEM algorithm below. As discussed the main difference to EM algorithm is the missigness weights introduced. 

\paragraph{The WEM Maximization step for exponential families.}
When $\P_{\rvx;\vtheta}$ is in an exponential family, the weighed maximization step has a simplified form. Let $T: \R^d\rightarrow \R^k$ denote $k$-dimensional sufficient statistics such that 
\[
\p_{\rvx}(\vx;\vtheta)=h(\vx)\exp[\vtheta^T T(\vx)-A(\vtheta)],
\] 
\cite{murphy2012machine}. Given a $\vphi$ and an estimate $\vtheta^{t-1}$, we define the weighted conditional sufficient statistic as
\[
\tilde{\T}(\mathcal{S})\defeq \sum_{i=1}^N \sum_{\vr\neq 0} \omega_{i,r}(\vphi) \E_{\rvx; \vtheta^{t-1}}[T(\vx) \mid \rvx_{o}=\vx_{o}^{(i)}].
\]
One can show that the maximization step of WEM (Eq. \ref{eq:wem}) i.e.

\[
	\vtheta^t = \argmax_{\vtheta} \sum_{i=1}^N \sum_{\vr\neq 0}\omega_{i,r}(\vphi)
	\E_{\rvx;\vtheta^{t-1}}\left[\log \P_{\rvx}(\rvx; \vtheta) \mid \rvx_{o}=\vx_{o}^{(i)}\right]
\]

is equivalent to 

\begin{equation}\label{eq:suff}
	\vtheta^t = \argmax_{\vtheta} \vtheta^T \tilde{T}(\mathcal{S})-N \cdot A(\vtheta)
\end{equation}

This optimization problem in Eq. \ref{eq:suff} has the same form for the Maximum Likelihood Estimation (MLE) with the compeltely observed data. The only difference lies in the sufficient statistics. Therefore, the same optimization procedure used for MLE estimation with fully observed data can be used with in maximization step of the WEM algorithm by replacing the original sufficient statistics with the weighted conditional sufficient statistics. This has been previously observed for the maximization step of the original EM algorithm\cite{dempster}. In our case we have an additional expectation with respect to the missing data patterns.


For the MVN distribution, the sufficient statistics is given by $\T(\vx)=(\vx,\vx \vx^T)^T$, and the update equations become:

\begin{equation}
	\begin{split}
		\mu_t &= \frac{\tilde{\T}(\data)_1}{N}=\frac{1}{N} \sum_{i=1}^N \sum_{\vr\neq 0} \omega_{i,r}(\vphi) \E_{\rvx; \vtheta^{t-1}}[\rvx \mid \rvx_{o}=\vx_{o}^{(i)}]\\
		\Sigma_t &= \frac{\tilde{\T}(\data)_2}{N}-\mu_t(\mu_t)^T\\
		&=	\frac{1}{N} \sum_{i=1}^N \sum_{\vr\neq 0} \omega_{i,r}(\vphi) \E_{\rvx; \vtheta^{t-1}}[\rvx \rvx^T \mid \rvx_{o}=\vx_{o}^{(i)}]-\mu_t(\mu_t)^T
	\end{split}
\end{equation}

As discussed, those equations are similar to EM update equations (See chapter 11.6.1.3 in \cite{murphy2012machine}) with the additional inner summation. The conditional expectations can be calculated using the Multivariate Gaussian update equations (See chapter 4.3.1 in \cite{murphy2012machine}).

\section{Experiment Details}

\subsection{Gaussian SCM, I}\label{app:scm1}
$\mB=\begin{bmatrix}
	0 & -.9 & -.8\\
	0 & 0 & 0\\
	0 & 0 & 0
\end{bmatrix}$, equal variance with $\sigma=1$. $N=1000$. 
\subsection{Gaussian SCM, II}\label{app:scm2}

We use equal variance with $\sigma=1$.

$\mB =\begin{bmatrix}
	0.  & -0.54& 0.  & 1.15& 0.  & 0. \\
	0.  & 0.  & 0.4& 0.  & 0.  & 0. \\
	0.  & 0.  & 0.  &-1.43&-0.9 & 1.29\\
	0.  & 0.  & 0.  & 0.  & 0.  & 0.  \\
	0.  & 0.  & 0.  & 0.  & 0.  & 0. \\
	0.  & 0.  & 0.  & 0.  & 0.  & 0. \\
\end{bmatrix}$

To reduce the number of missing data we used a variation of the local rejection sampling Lem \ref{lem:localized_generalized_rejection_sampling}. Specificially, when a sample is accepted instead of uniformly distribution missing data patttern, we constrained it to be all or none i.e. 

\begin{equation}
	\P_{\rvr|\rvx}(\vr|\vx) = \begin{cases}
		 \frac{\Lambda(\vx_{\mathcal{V}})}{\Lambda} & \text{ if } \vr_{\overline{\mathcal{V}}}=1 \text{ and } \vr_{\mathcal{V}} = 1 \\
		1 - \frac{\Lambda(\vx_{\mathcal{V}})}{\Lambda} & \text{ if } \vr_{\overline{\mathcal{V}}}=1 \text{ and } \vr_{\mathcal{V}} = 0 \\
		0 & \text{ otherwise}
	\end{cases}.
\end{equation}

In addition, the maximum density ratio $\Lambda$ is selected as maximum density ratio over the observed samples.

\subsection{Setup for \namenn } \label{app:nn}

 We have used Tensorflow v2.4. \cite{tensorflow} for implementation. We have implemented a Keras model with a custom training loop which includes the WEM Algorithm (Alg. \ref{alg:wem}) as subroutine. In our experiments, we have used $K=5$, random initializations for the \namenn which are sampled using \text{make\_spd\_matrix} function of \cite{scikit-learn} and scaled such that the diagonal entries match the empirical variances. We used $\epsilon=1e-5$ for the stopping criterion of WEM. 
 
 \subsection{NN parameters}
 
 \begin{itemize}
 	\item Weight initialization: Tensorflow default
 	\item Custom loss function as defined in Algorithm \ref{alg:nn}:
 	\[\sum_{k=1}^K\ell (\ttil_k, \ta,\vphi,\lambda)\]
 	\item Stopping criterion
 	\begin{enumerate}
 		\item Sachs: Fixed 300 epochs
 		\item Gaussian SCM,I: early stopping (restore best, min change = 1e-4, patience =10,monitor=loss function of the final initialization in WEM)
 		 	\end{enumerate}
 		\item Optimizer: Adam, 	learning rate = 1e-3 (Sachs ), 1e-2 (Gaussian SCM,I) 

 \end{itemize}

\subsection{Hyper-parameters for the Causal Discovery Algorithms}
\subsubsection{NOTEARS}
We used the code\footnote{https://github.com/xunzheng/notears/tree/master/notears} provided by the authors of \cite{notears} and used didn't change the hyper-parameters i.e. 

\begin{itemize}
	\item \text{max\_iter}=100, 
	\item \text{h\_tol}=1e-8, 
	\item \text{rho\_max}=1e+16, 
	\item \text{w\_threshold}=0.
\end{itemize}
\subsubsection{PC, missPC}
We used package causal-learn available at \footnote{https://github.com/py-why/causal-learn}. We used the hyper-parameters used an example test file:
For PC:
\begin{itemize}
	\item \text{pc\_alpha}=0.01
	\item \text{indep\_test}=``fisherz''
	\item \text{uc\_priority}=-1.
\end{itemize} 

For missPC:
\begin{itemize}
	\item \text{pc\_alpha}=0.01
	\item \text{indep\_test}=``mv\_fisherz''
	\item \text{uc\_priority}=-1.
	\item mvpc=True
\end{itemize} 

\subsubsection{missDAG}\label{app:init}
Algorithm is provided in Alg. \ref{alg:em}.

\begin{itemize}
	\item Initialization: Five different initialization schemes for the covariance matrix: using a diagonal matrix scale to match the estimated variances (``Emp. Diag.''),
	using an identity matrix (``Ident.''), the ground truth covariance matrix (``True''), a scaled random covariance matrix Random(*)\footnote{Generated using the \texttt{make\_spd\_matrix} function of scikit-learn and scaled to match the empirical variances.}, and a scaled inverse Wishart random matrix IW(*) \footnote{It uses the \texttt{make\_spd\_matrix} function of scikit-learn for the prior parameter.}.	
	\item Variance equality: equal variance (Gaussian SCM I, and II), non-equal variance (Sachs dataset)
	\item  Stopping criterion: $\epsilon=1e-5$.
\end{itemize}

\subsection{MissDAG Algorithm}

We have derived our equations and implemented our version based on the description in \cite{missdag} as their code was unavailable.

MissDAG uses the EM algorithm but optimizes the maximization step with a DAG constraint. Let N i.i.d. samples from $P_{X;\tr}$ and $P_{R\mid X}$ are denoted by $x^{(1)},\dots, x^{(N)}$ and $R^{(1)},\dots, R^{(N)}$ respectively. We denote $o^{i}=\{j:R^{(i)}_j=1\}$ and its complement as $-o^{i}$. Following Eq. xx \cite{missdag}, given $\vtheta^{t-1}$ (parameters from the previous step),  the maximization step of the EM algorithm is

\begin{equation}\label{eq:em}
	\vtheta^t = \argmax_{\vtheta\in \mathcal{D}} \sum_{i=1}^N 
	\E_{\rvx;\vtheta^{t-1}}\left[\log \P_{\rvx}(\rvx; \vtheta) \mid \rvx_{o^i}=\vx_{o^i}^{(i)}\right]
\end{equation}

 Under the zero mean-Gaussian SCM with equal variance noise assumption the likelihood has a simple form (see \cite{golem} appendix ) and because Gaussian SCM is an exponential family distribution the Eq. \ref{eq:em} also has a simple form. Let $\hat{\mathcal{S}}=\{\vx_{o^1}^{(1)},\dots,\vx_{o^N}^{(N)}\}$, we denote the aggregate sufficient statistics as

\begin{equation}\label{eq:suffhat}
	\hat{\T}(\hat{\mathcal{S}})\defeq \sum_{i=1}^N \E_{\rvx; \vtheta^{t-1}}[\rvx\rvx^T \mid \rvx_{o^i}=\vx_{o^i}^{(i)}]
\end{equation}
Given $W\in \real^{d\times d}$, let $W_j$ denotes the j'th column and similarly $\hat{\T}(\hat{\mathcal{S}})_j$ denotes the j'th column of the aggregate sufficient statistic. It can be shown that $Eq. \ref{eq:em}$ is equals to 

\begin{equation} \label{eq:missdag}
	\begin{split}
		W^t &= \argmin_{\mB\in \mathcal{D}} \sum_{j=1}^d  \hat{\T}(\hat{\mathcal{S}})_{j,j}-2\inner{\mB_j}{\hat{\T}(\hat{\mathcal{S}})_j}+ \inner{\mB_j}{\hat{\T}(\hat{\mathcal{S}})\mB_j} \\
	\end{split}
\end{equation}

For $\sigma$ is given under the equal variance assumption as 
\begin{equation}\label{eq:miss1}
			\sigma^t = \dfrac{1}{Nd}\sum_{j=1}^d  \hat{\T}(\hat{\mathcal{S}})_{j,j}-2\inner{\mB_j^t}{\hat{\T}(\hat{\mathcal{S}})_j}+ \inner{\mB_j^t}{\hat{\T}(\hat{\mathcal{S}})\mB_j^t}
\end{equation}
Under the non-equal variance assumption it is 
\begin{equation}\label{eq:miss2}
	\sigma^t_j = \dfrac{1}{N}\hat{\T}(\hat{\mathcal{S}})_{j,j}-2\inner{\mB_j^t}{\hat{\T}(\hat{\mathcal{S}})_j}+ \inner{\mB_j^t}{\hat{\T}(\hat{\mathcal{S}})\mB_j^t}
\end{equation}

Following \cite{missdag} we used the NOTEARS package to solve Eq. \ref{eq:missdag}. We implemented Eq. \ref{eq:missdag} as a custom loss function and provided its gradient with the python package Autograd \cite{autograd}. The empirical likelihood is given by:

\[\hat{J}(\theta)=\sum_{i=1}^N\log \P_{\rvx_{o^i}}(\vx_{o^i}^{(i)}; \vtheta) \]

The overall missDAG algorithm is given as follows:

\begin{algorithm}[H]
	\caption{Our implementation of the MissDAG Algorithm \cite{missdag}. }
	\label{alg:em}
	
	\begin{algorithmic}
		\STATE {\bfseries Input: $ \epsilon, \hat{\data}$ }
		\STATE $t\leftarrow 1$
		\WHILE{$\hat{J}(\vtheta^{(t)})-\hat{J}(\vtheta^{(t-1)})\geq \epsilon |J(\vtheta^{(t-1)},\bm{\phi})|$} 
		\STATE $\hat{\T} \leftarrow $ Eq. \ref{eq:suffhat} using  $\vtheta^{(t-1)},\hat{\mathcal{S}}$
		\STATE $\mB^t \leftarrow $ Eq. \ref{eq:missdag} via NOTEARS using $\hat{\T}$
		\STATE $\sigma^t \leftarrow $ Eq. \ref{eq:miss1} or Eq. \ref{eq:miss2} using $\mB^t,\hat{\T}$
		\STATE $\Sigma^t \leftarrow (I-\mB)^{-1}D(\sigma^t)(I-\mB)^{-T}$
		\STATE $\vtheta^t=\Sigma^t$
		\STATE $t\leftarrow t+1$
		\ENDWHILE
		\STATE $\that \leftarrow \vtheta^{t-1}$
		\STATE {\bfseries Output: $\that$} 
	\end{algorithmic}
\end{algorithm}

\section{Additional Results for the Sachs Dataset}\label{app:sachs}

\begin{figure}[H]
	\centering
	\includegraphics[width=1\columnwidth]{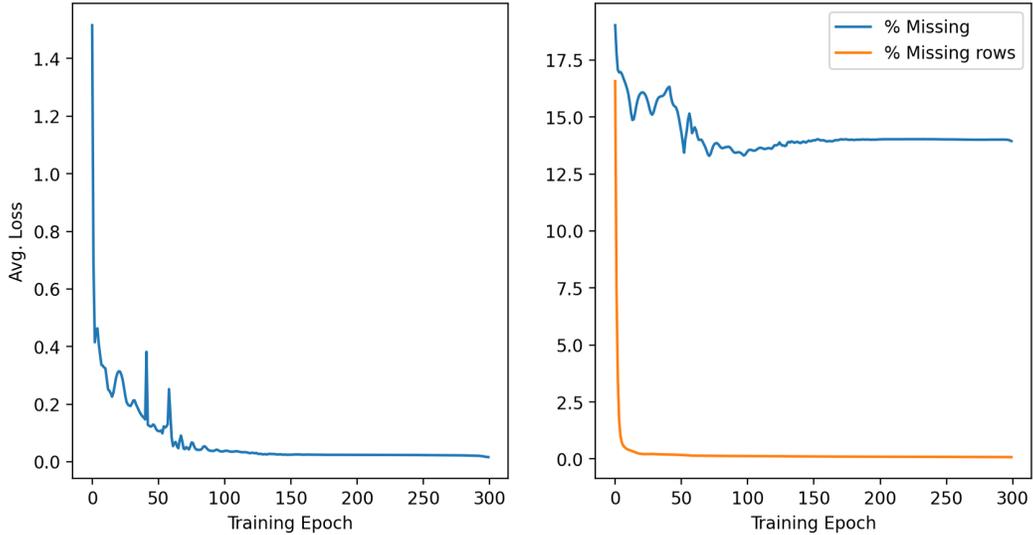}
	\caption{Model training curves for \namenn. Loss function over epochs (left). $\%$ missingness and $\%$ fully missing rows in the masked columns (in $\midx$).}
	\label{fig:loss}
\end{figure}
\begin{figure*}
	\begin{subfigure}[\namenn]{}
		\includegraphics[width=1.\textwidth]{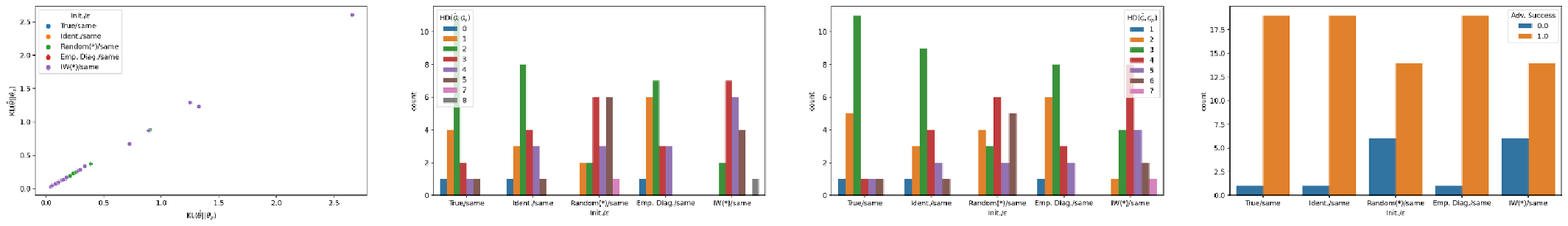}
	\end{subfigure}
	\begin{subfigure}[MCAR]{}
		\includegraphics[width=1.\textwidth]{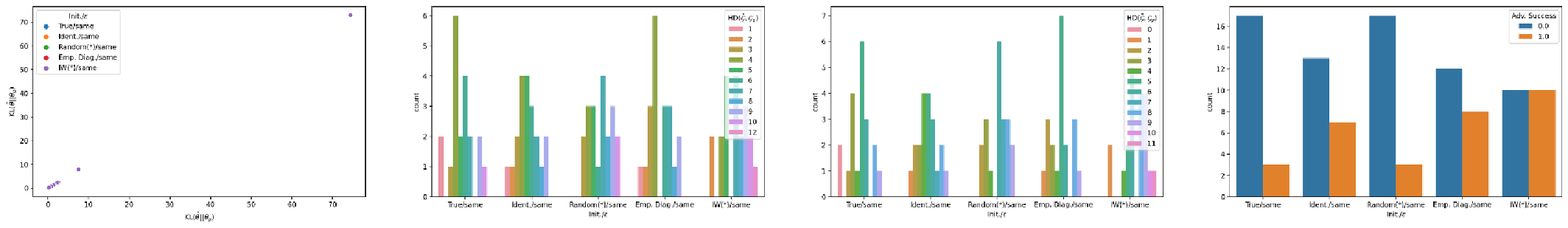}
	\end{subfigure}
	\caption{Results for Sachs dataset. Each scatter plot is a different initialization. Left to right: $\text{D}_{\text{KL}}(\P_{\rvx;\that} \| \P_{\rvx;\ta})$ vs $\text{D}_{\text{KL}}(\P_{\rvx;\that} \| \P_{\rvx;\tr})$; distribution of HD($\hat{\mathcal{G}},\mathcal{G}_{p}$); distribution of HD($\hat{\mathcal{G}},\ga$); adversarial success.}
\end{figure*}

\end{document}